\documentclass{article}

\usepackage{microtype}
\usepackage{graphicx}
\usepackage{subfigure}
\usepackage{booktabs} 

\usepackage{hyperref}

\usepackage{amsmath,amsfonts,bm}

\def\eqref#1{equation~\ref{#1}}

\def\1{\bm{1}}

\def\rmX{{\mathbf{X}}}

\DeclareMathAlphabet{\mathsfit}{\encodingdefault}{\sfdefault}{m}{sl}
\SetMathAlphabet{\mathsfit}{bold}{\encodingdefault}{\sfdefault}{bx}{n}

\usepackage[accepted]{mlsys2023}

\usepackage[utf8]{inputenc} 
\usepackage[T1]{fontenc}    
\usepackage{url}            
\usepackage{booktabs}       
\usepackage{amsfonts}       
\usepackage{nicefrac}       
\usepackage{microtype}      
\usepackage{graphicx}

\usepackage[ruled,vlined]{algorithm2e}

\usepackage{microtype}
\usepackage{graphicx}
\usepackage{subfigure}
\usepackage{enumitem}
\usepackage{pifont}
\usepackage{adjustbox}
\usepackage{caption}
\usepackage{booktabs} 
\usepackage{color, colortbl}
\usepackage[dvipsnames]{xcolor}
\usepackage{footnote}
\usepackage{multirow}
\usepackage{listings}

\usepackage{graphicx}
\usepackage{amsmath}
\usepackage{subfigure}
\usepackage{tcolorbox}
\usepackage{multirow}
\usepackage{pifont}

\usepackage{amssymb}
\usepackage{diagbox}

\newcommand{\eg}{\textit{e.g.}}
\newcommand{\ie}{\textit{i.e.}}
\newcommand{\etc}{\textit{etc}}

\def\equationautorefname~#1\null{Eq.~(#1)\null}

\mlsystitlerunning{SysNoise: Exploring and Benchmarking Training-Deployment System Inconsistency}

\begin{document}

\twocolumn[
\mlsystitle{SysNoise: Exploring and Benchmarking Training-Deployment System Inconsistency}



\mlsyssetsymbol{equal}{*}

\begin{mlsysauthorlist}
\mlsysauthor{Yan Wang}{equal,st}
\mlsysauthor{Yuhang Li}{equal,st,ya}
\mlsysauthor{Ruihao Gong}{equal,st,bh}
\mlsysauthor{Aishan Liu}{equal,bh}
\mlsysauthor{Yanfei Wang}{st}
\mlsysauthor{Jian Hu}{st}
\mlsysauthor{Yongqiang Yao}{st}
\mlsysauthor{Yunchen Zhang}{st}
\mlsysauthor{Tianzi Xiao}{st}
\mlsysauthor{Fengwei Yu}{st}
\mlsysauthor{Xianglong Liu}{bh}
\end{mlsysauthorlist}

\mlsysaffiliation{st}{SenseTime Research, Beijing, China}
\mlsysaffiliation{bh}{Beihang University, Beijing, China}
\mlsysaffiliation{ya}{Yale University, New Haven, USA}

\mlsyscorrespondingauthor{Yan Wang}{mrwangyan98@gmail.com}

\mlsyskeywords{Machine Learning, MLSys}


\begin{abstract}
Extensive studies have shown that deep learning models are vulnerable to adversarial and natural noises, yet little is known about model robustness on noises caused by different system implementations. In this paper, we for the first time introduce SysNoise, a frequently occurred but often overlooked noise in the deep learning training-deployment cycle. In particular, SysNoise happens when the source training system switches to a disparate target system in deployments, where various tiny system mismatch adds up to a non-negligible difference. We first identify and classify SysNoise into three categories based on the inference stage; we then build a holistic benchmark to quantitatively measure the impact of SysNoise on 20+ models, comprehending image classification, object detection, instance segmentation and natural language processing tasks. Our extensive experiments revealed that SysNoise could bring certain impacts on model robustness across different tasks and common mitigations like data augmentation and adversarial training show limited effects on it. Together, our findings open a new research topic and we hope this work will raise research attention to deep learning deployment systems accounting for model performance. We have open-sourced the benchmark and framework at \url{https://modeltc.github.io/systemnoise_web}.
\end{abstract}
]



\printAffiliationsAndNotice{\mlsysEqualContribution} 

\section{Introduction}

Deep neural networks have demonstrated remarkable success in handling multiple tasks~\cite{AlexNet,VGG,ResNet,devlin2018bert,GPT3}, yet they are vulnerable against noises. Despite the progress devoted to noises made by human-being or nature  (\eg, adversarial noises \cite{goodfellow2014explaining} and natural noises~\cite{hendrycks2019robustness}), little is known about model robustness on noises caused by different system implementations. In practice, the model deployment often faces diverse implementation platforms spanning from general (\eg, CPU, GPU) to specialized (\eg, NPU, ASIC) computing hardware; from the cloud server to edge devices; and often with different back-ends (\eg, TensorRT~\cite{tensorrt} for GPUs, SNPE~\cite{snpe} for DSPs, CANN~\cite{cann} for Ascend). These different software-hardware system implementations would bring certain noises resulting in considerable model performance degeneration. More importantly, these noises cannot be completely prohibited as long as a trained model will be deployed to multiple target platforms.

\begin{figure*}[t]

    \centering
    \includegraphics[width=1.0\linewidth]{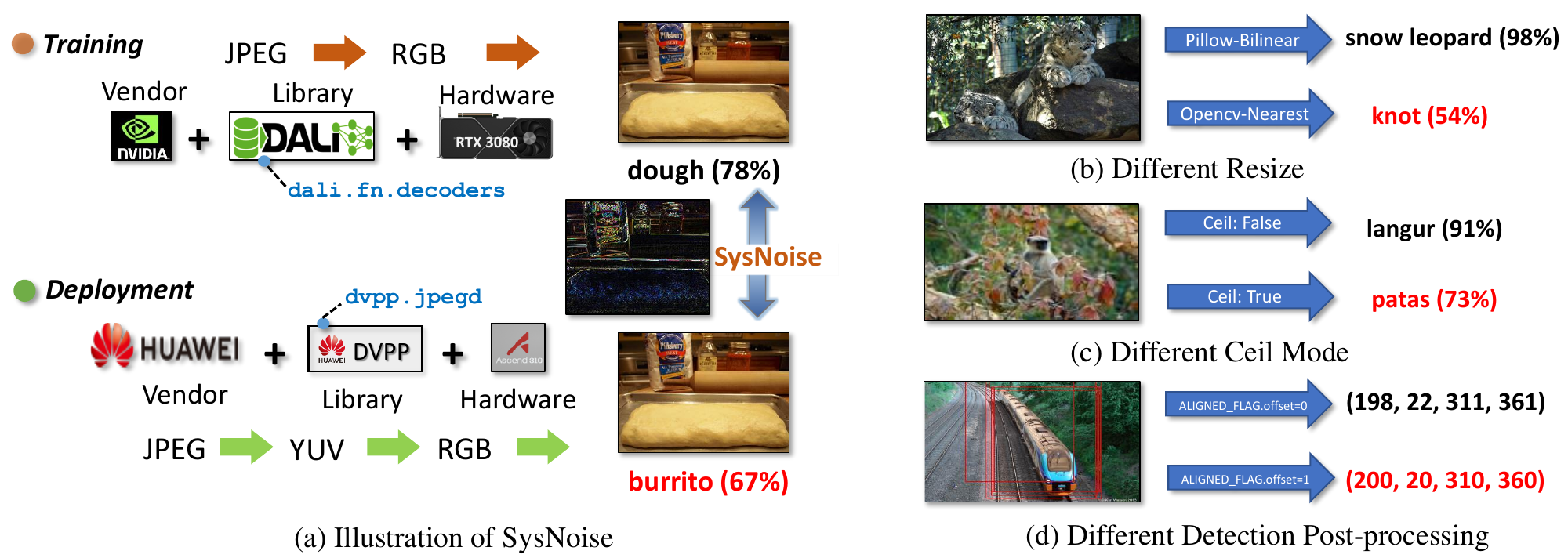}
    \caption{An illustration of SysNoise (a) and its negative effect on model robustness (b-d). Here we take noises from the decoder as an example. We usually use the DALI library from NVIDIA on GPU during training and the DVPP library from HUAWEI on Ascend during deployment for decoding acceleration, which results in minor decoding differences and would mislead the prediction.}
    \label{fig_frontpage}
        
\end{figure*}

Thus, in this paper, we pioneeringly discuss an unwanted yet non-negligible type of noise caused by the inconsistency of the training-deployment system (see \autoref{fig_frontpage} for illustration), deemed as system noise~(\textit{abbrev. SysNoise}). Based on where SysNoise could happen, we classify it into three different types. \ding{192}~\textit{Pre-processing:} Depends on the implementation of input data. For example, different image decoding (JPEG2RGB) algorithms and different interpolation methods for image resize and crop. 
\ding{193}~\textit{Model Inference:} Caused by different implementations of the model during inference. For instance, models with the same parameters can have different results when the upsampling operator is different. Using different data types (INT8, FP16, FP32) also leads to different accuracy. \ding{194}~\textit{Post-processing:} Includes the further manipulation of inference results, \eg, applying softmax function in classification tasks and calculating the bounding box in detection tasks. Overall, SysNoise exhibits its impact on the whole inference pipeline, leading to an undesired performance drop.

To better understand and comprehensively evaluate the influence of SysNoise on the deployed model, we provide a thorough quantitative benchmark on 3 common computer vision tasks (\ie, classification, detection, and segmentation) with 20+ representative models and typical baselines. As for natural language processing, we provide a benchmark on OPT~\cite{opt} model on 4 datasets. Our large-scale experiments reveal several insights: (1) though these noises are not chosen by any adversary, SysNoise would bring considerable impacts on model robustness, and could cause up to $9.97\%$ and $10.67\%$ drops on classification and detection tasks respectively; (2) different architecture families induce different robustness on SysNoises (\eg, ViTs and CNNs), even in the same architecture family, a larger model tends to have low variance and low accuracy degradation on SysNoise; and (3) SysNoise seems to be highly diverse and different from adversarial and natural noises, where common mitigations like data augmentation and adversarial training show limited effects on it. Together with existing benchmarks on adversarial and natural noises, we could build a more comprehensive and general understanding and ecosystems for robustness benchmarking involving more perspectives. This benchmark for evaluating robustness to system noises provides useful information, and hopefully, it can open a new research direction for building robust deep learning deployment systems. 

In conclusion, our contributions can be summarized as threefold:
\begin{enumerate}[nosep, leftmargin=*]
\item For the first time, we identify and systematic research on an important problem named SysNoise (ranging from pre-processing, model inference, and post-processing noise), which is caused by the training-deployments system inconsistency. 
\item We build a benchmark and framework to quantitatively evaluate SysNoise on 20+ deep neural networks, including image classification (ImageNet), detection (MS COCO), segmentation (CitySpace), and natural language processing. 

\item We conducted in-depth analyses and found several insights, which revealed that SysNoise is an inevitable and urgent-to-solve problem for both algorithm researchers and hardware vendors. 
\end{enumerate}

\section{Related Work}
\textbf{Noises Types and Benchmarks.} Extensive shreds of evidence have shown that deep learning models are unstable towards different noises, including adversarial noises and natural noises. \emph{Adversarial noises}, which are imperceptible to human vision, could easily make neural networks misclassify the input images~\cite{szegedy2013intriguing,fgsm,liu2022harnessing,liu2023x,Liu2019Perceptual,liang2021parallel,Wang_2021_CVPR,liu2020bias,liu2020spatiotemporal}. To benchmark and evaluate adversarial robustness, \cite{su2018is} first investigated the adversarial robustness of 18 models on ImageNet; \cite{Ling2019Deepsec} built the  platform DEEPSEC for adversarial robustness analysis including 16 adversarial attacks, and 13 adversarial defenses; meanwhile, RealSafe \cite{Dong2020Benchmarking} open-sourced and benchmarked adversarial robustness on image classification tasks. More recently, large-scale benchmarks on adversarial robustness regarding defense strategies (RobustBench \cite{croce2020robustbench}) and model architectures (RobustART \cite{tang2021robustart,liu2023exploring}) were developed. Besides adversarial noises, there exist another type of model-agnostic noise named \emph{natural noises} (also deemed as corruptions), which are commonly witnessed in the real-world scenario, \eg, blur, snow, and frost. Some representative datasets are constructed to simulate and benchmark the natural noises, such as ImageNet-P, ImageNet-C~\cite{hendrycks2019robustness}, and ImageNet-A, ImageNet-O~\cite{hendrycks2021nae}. \cite{hendrycks2020many} also introduced new real-world distribution shift datasets including changes in image style, geographic location \etc. However, these studies only focus on noises brought during data acquisition, while ignoring the impacts of the whole inference pipeline caused by different system implementations.

Furthermore, some studies introduce the influence of individual SysNoise. \cite{yan2021real, boltaevich2019estimationresize} show how image pre-processing progresses including image decoder, resize method, and color conversion generate noise. However, they only introduce one or two noises in image pre-processing and lack investigation on the whole training-deployment progress as well as the combination of system noise.
\cite{biterror} introduces the bit error that is caused by the Low-voltage operation of DNN accelerators, which does illustrate that training-deployment system inconsistency can bring error. And \cite{randomnessintraining} show the random noise caused by different training systems. But this work only focuses on differences in the training system and ignores the deployment system.
In addition, ~\cite{jia2021exploiting} takes the first step towards the influence of the floating-point value representation. They highlight that, to achieve practically reliable verification of neural networks, the system must accurately model the effects of any floating-point computations. However, this paper only conducts a preliminary attempt at the effect of floating-point numerical error for neural network verifiers. 


\textbf{Approaches to Improving Model Robustness.} To improve model robustness against \emph{adversarial noises}, a long line of adversarial defense works have been proposed including: (1) adversarial training that adversarially train deep models using adversarial examples \cite{goodfellow2014explaining,PGD,tramer2017ensemble,shafahi2019adversarial,liu2021ANP,zhang2021interpreting}; and (2) adversarial detection that distinguishes the clean example and adversarial example \cite{grosse2017statistical,gong2017adversarial,jiang2020attack}. 

To effectively tackle the \emph{natural noises}, several studies have been devoted primarily from the perspective of data augmentation. By producing an elementwise convex combination of two images, Mixup~\cite{mixup2017} could regularize neural networks to favor simple linear behavior in-between training examples and improve model performance. Different from Mixup, AutoAugment \cite{2018AutoAugment} adopts and tunes a group of augmentations to optimize performance on a downstream task. To further improve model robustness against natural noises, AugMix \cite{hendrycks2020augmix} was proposed to mix multiple augmented images. And APR-SP~\cite{chen2021amplitude} was proposed to force the CNN to pay more attention on the structured information from phase components and keep robust to the variation of the amplitude which can help with the model's robustness of natural noise.

Test-time adaptation is another way to improve the model's performance at inference. It refers to adapting a machine learning model to a target domain at test time, without access to the source data or even any additional labeled/unlabeled samples from the target distribution to fine-tune the source model. \cite{tent} propose a method to reduce generalization error by reducing the entropy of model predictions on test data, and it reduces error for image classification on corrupted ImageNet and CIFAR-10/100 and reaches a new state-of-the-art error on ImageNet-C.



\begin{figure*}[t]
    \centering
    \includegraphics[width=\linewidth]{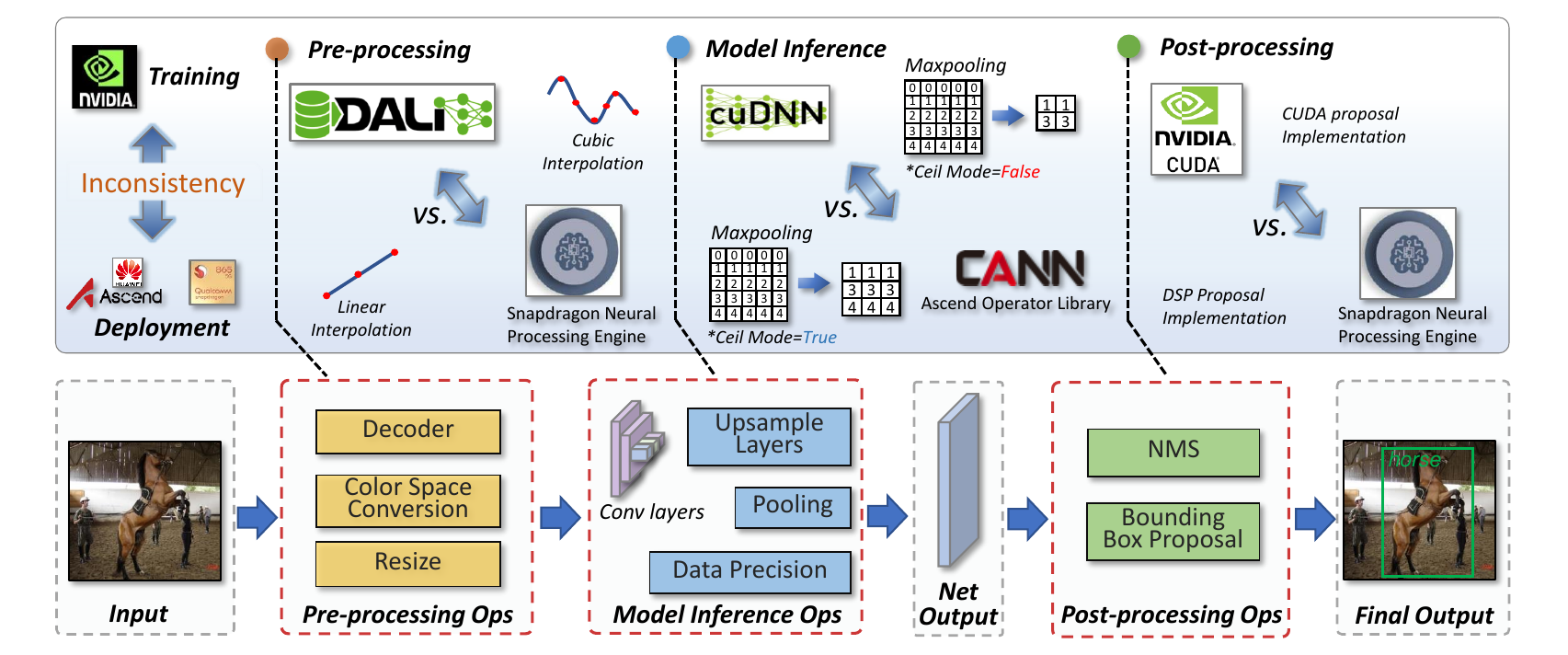}
    \caption{\textbf{Overview of SysNoise}. SysNoise is caused by an inconsistent implementation between the training system and deployment system, consisting of three parts, namely pre-processing noise, model inference noise, and post-processing noise. }
    \label{fig_overview}
\end{figure*}

\section{System Noise Benchmark}

In this section, we introduce the benchmark for system noise. First, we summarize the three stages in SysNoise, namely pre-processing noise, model inference noise, and post-processing noise as shown in~\autoref{fig_overview}. Then, we introduce these three stages SysNoise in detail. Note that we only give the basic principles, a more rigorous mathematical difference of SysNoise is provided in \autoref{appendix_math}. 

\subsection{Pre-processing Noise}
\label{sec_prep}
Pre-processing means the preparation of the input tensor of the neural network. Concretely, in computer vision tasks, the pre-processing will convert an image raw file to a 3-dimension tensor (width, height, and RGB channels). To fulfill this conversion, two steps are required. First, the raw file (JEPG) will be decoded to a tensor with the image's original shape. Then, the tensor will be resized to a certain shape. 
To decode the image from the JEPG file to an RGB tensor, it is required to perform the inverse Discrete Cosine Transform (iDCT) operation. 
In theory, the principle of iDCT is fixed, but we find decoding one image file in different third-party libraries (\eg, OpenCV~\cite{opencv_library}, Pillow~\cite{umesh2012image}, FFmpeg~\cite{tomar2006converting}) will output different RGB tensors. This is because some libraries prefer to use Fast iDCT~\cite{chen1977fast} instead of the vanilla one, which may sacrifice the image quality for the decoding speed. 
Furthermore, there would be some minor errors in the decoding implementation, such as the cosine function. These minor errors can cause a shift in the pixel values of the final RGB image tensor.
As a result, when changing the decoding tools used in training to another one in inference, we observe a drop in accuracy.


The second cause of pre-processing noise is image resize. Resize is a simple scaling operation that adjusts resolution to a different size, either up (increase resolution) or down~(decrease resolution). In a resize operation, one needs to predict the pixel value at an unseen position. This is often performed by different \textit{interpolation} algorithms. 
For example, nearest-neighbor interpolation directly selects the value of its nearest know pixel. While bilinear interpolation predicts the unknown pixel by computing the distance-based weight average of the existing neighbor 4 pixels, \ie~top, bottom, left, and right, which has a rather continuous interpolation effect. 
Besides, there are many other interpolation methods. In Appendix, we provide the detailed mathematical explanation of these interpolation algorithms as well as the supporting package in computer vision. Note that the difference in interpolation may even occur at the package level, \ie~even the same interpolation algorithm might differ in different supporting packages.

The third source of inconsistency during the pre-processing stage comes from the conversion of color space. In the practical application, there are various representation formats for videos and images, \eg, RGB and YUV. The RGB format defines the color space with the value of red, green, and blue channels while the YUV format separates the brightness information (Y) from the color information (U and V), which is the format native to TV broadcast and composite video signals. To save the required storage, different variants of the YUV format are devised. Among them, the NV12 format can encode one pixel with only 12bits, enjoying a low memory consumption and high efficiency. Therefore, many decoder accelerators such as Microsoft DirectX Video Acceleration and Ascend 310 adopt this format. However, for the training of most neural networks, input images are fed with the RGB format. Decoding images to YUV and then converting it to RGB is difficult to output the same direct RGB decoded images.



\newcommand{\mypm}[1]{{~(#1)}}

\subsection{Model Inference Noise}
Model inference noise accounts for the difference that happens during the inference process. This is primarily due to the implementation of various operations. For example, the convolution can be implemented in many ways (GEMM, Img2Col, Winograd, etc). 
We primarily discover 3 three types of model inference noise that cause a performance drop. 
The first one is the ceiling mode for max-pooling layers. Ceiling mode means how to compute the output spatial shape. Setting ceiling mode to true will allow the sliding windows to go off-bounds if they start within the left padding. Hardware vendors usually support different ceiling modes, causing an inevitable mismatch. 

Another important type of model inference noise is the upsampling method. It is widely used in segmentation task. And in the detection task, the widely used feature pyramid networks~\cite{lin2017feature} integrate the features from different stages within the network. These features have an uneven resolution, requiring an upsampling operation to match feature resolution. Same as the resize operation we discussed in~\autoref{sec_prep}, the choice of the interpolation in upsampling layers can play an important role and lead to different predictions. We find that the FPN is quite sensitive to interpolation. 

Finally, the precision of data representation can also be viewed as a type of model inference noise. Generally, the input data and the parameters in the model are stored with 32-bit floating-point numbers. 
However, some hardware systems may restrict the precision, \eg, only 16-bit floating-point numbers or 8-bit integers are allowed. 
Low-bit numbers unavoidably preserve less information than the full-precision numbers, causing accuracy degradation. 
Note that in the field of quantization research, some training methods could alleviate this problem~\cite{jacob2018quantization}. We do not use such a training-compensated method here, in order to evaluate how much the deep learning model can resist under low data precision and how a single type interacts with other types of SysNoise, even though there are contingency methods.

\subsection{Post-Processing Noise}

Post-processing is used to convert the network output to the prediction results. In image classification, this refers to the $\mathrm{Softmax}$ function which applies the exponential function to normalize the output to $(0, 1)$. In object detection, the predicted output of the network needs to be calculated to the final bounding box. During this process, there are rounding operations to get integer resolution coordinates. Then, all the candidate bounding boxes will be sorted with the predicted confidence and filtered with non-maximum suppression. This procedure is easy to introduce noises in detail, \eg, the rounding up or rounding down choice, \etc. Many hardware vendors provide black-box implementations of these operations to accelerate the deployment. Unfortunately, we find that they often fail to produce the same results, causing an impact on the final performance.

\subsection{Benchmarking SysNoise}
\label{benchmark_sysnoise}



\newcommand{\cmark}{\color{ForestGreen}\ding{51}}%
\newcommand{\xmark}{\color{Red}\ding{55}}%
\newcommand{\clsdet}{$\mathrm{Cls/Det/Seg}$}
\begin{table*}[htbp]
\caption{\textbf{List of our discerned system noise,} including 3 stages (pre-processing, model inference, post-processing). Affected tasks consists of image classification~($\mathrm{Cls}$), object detection~($\mathrm{Det}$), semantic segmentation~($\mathrm{Seg}$) } and natural language processing~($\mathrm{NLP}$)
\label{tab_sysnoise_type}
\centering
\begin{adjustbox}{max width=\textwidth}
\begin{tabular}{lcccccccc}
\toprule
\textbf{Stage} & \multicolumn{3}{c}{\textbf{Pre-processing}} & \multicolumn{3}{c}{\textbf{Model inference}} & \multicolumn{1}{c}{\textbf{Post-processing}} \\
\cmidrule(l{2pt}r{2pt}){2-4}\cmidrule(l{2pt}r{2pt}){5-7}
\cmidrule(l{2pt}r{2pt}){8-8}
Type & Decoder & Resize & Color Space & Ceil Mode & Upsample & Data Prec. & Detection Proposal \\
\midrule
Task & \clsdet & \clsdet & \clsdet & \clsdet & $\mathrm{Det/Seg}$ & $\mathrm{Cls/Det/Seg/NLP}$ & $\mathrm{Det}$ \\ Input Dependence & \xmark & \xmark &  \cmark & \xmark &  \xmark & \cmark &  \xmark \\
Noise Effect Level & High & Very High & Middle & High & Very High & High & Middle \\
Number of Categories & 4 & 11 & 2 & 2 & 2 & 3 & 2 \\
Occurrence Frequency & Very High & Very High & High & High & Middle & High & Middle \\
\bottomrule
\end{tabular}
\end{adjustbox}
\end{table*}


\textbf{Types of SysNoise.} SysNoise originates from the implementation difference in hardware and software. In \autoref{tab_sysnoise_type}, we briefly summarize the types of SysNoise in each stage, as well as their applied task, dependence on input data, level of effect, and the number of categories. We highlight that here we view SysNoise as random noise since in practice it is inflexible to train a unique model for corresponding hardware.

$\bullet$ \emph{Preprocessing Noise}
    \begin{enumerate}[nosep, leftmargin=1.9em]
        \item Decoder: To simulate noise during decoding process, four different python packages are selected to decode images — PIL~\cite{umesh2012image}, OpenCV~\cite{opencv_library}, FFmpeg ~\cite{ffmpeg_library} and DALI~\cite{nvidia-dali}, which implement their own image decode function, and output different image tensors. 
        \item Resize: We choose up to 11 different resize methods to represent noise that occurred in image resizing. Specifically, we utilize two Python packages, the Pillow and the OpenCV. For Pillow, we adopt interpolations from \{bilinear, nearest, box, hamming, bicubic, lanczos\} methods, and for OpenCV, we adopt interpolations from \{bilinear, nearest, area, bicubic, lanczos\}. 
        \item Color mode: To simulate noise that comes from the conversion of color space, we generate the noised images by first decoding the images to RGB and then transforming them to YUV color space and then back to RGB with Ascend Computing Language~(ACL)~\cite{cann}.
    \end{enumerate}
    
$\bullet$ \emph{Model Inference Noise}
    \begin{enumerate}[nosep, leftmargin=1.9em]
        \item Ceil mode: This can only be tested on models which has stride 2 max-pooling layers, such as ResNets \cite{he2016deep}. We train the model with floor mode but test it with ceil mode.
        \item Upsample: Nearest neighbor and bilinear are the two most commonly supported algorithms for upsampling. Following~\cite{faster-fpn}, we train the original upsample layers with nearest-neighbor interpolation and test it with bilinear interpolations.
        \item Data Precision: To evaluate the model's robustness under different precisions, we quantize the model to FP16 or INT8 and test it. 
    \end{enumerate}
    
$\bullet$ \emph{Postprocessing Noise}
    \begin{enumerate}[nosep, leftmargin=1.9em]
        \item Detection proposal: We evaluate the influence of whether to add the value of 1 when calculating bounding boxes from offsets, both of which are common in hardware implementations.
\end{enumerate}

\textbf{Evaluation Metrics.} For classification/detection/segmentation/natural language processing, we report the top-1 accuracy/mean Average Precision/mean Intersection over Union difference for measuring the robustness of models. If the SysNoise has multiple options, we report the mean difference as well as the max difference, otherwise, only the metric difference is reported.

\section{Experiment and Analysis}

In this section, we conduct a thorough benchmark and analysis on the SysNoise.
In~\autoref{sec_cls}, we illustrate the experimental setting for image classification, detection, segmentation, and NLP tasks; in~\autoref{sec_results} we extensively evaluate all types of SysNoise on these four tasks; in~\autoref{sec_interp}, we interpret the SysNoise by comparing it with natural noise and adversarial noise as well as some visualizations. 

\subsection{Experimental Setting}
\textbf{Classification Task.}
\label{sec_cls}
We benchmark SysNoise on the ImageNet dataset for the classification task, including both Convolutional Neural Networks (CNNs) and Vision Transformers (ViTs). 
For CNNs, we evaluate ResNet~\cite{he2016deep}, MobileNetV2~\cite{sandler2018mobilenetv2}, RegNet~\cite{radosavovic2020designing}, and EfficientNet~\cite{tan2020efficientnet} families. 
In addition, we evaluate an extremely small architecture — MCUNet~\cite{lin2020mcunet}, which only has 0.74MB parameters. 
For ViTs we evaluate the original Vision Transformer~\cite{dosovitskiy2020vit} and the Swin Transformer~\cite{liu2021Swin} families.
Each family covers different computation and memory budgets to ensure both large and tiny models are verified. During training, we use Nvidia DALI ~\cite{dali} to prepare data, \ie, image decode, resize and color space are configured by default function in DALI. All models take an input shape of $224\times224$ except EfficientNet. We train the default model using FP32 format as this is the standard format in GPU training. For ResNet, we train it with the floor mode of its max-pooling layer. All other training settings follow the original settings of the model.

\textbf{Detection and Segmentation Task.}
\label{sec_detseg}
For object detection, we use COCO dataset and adopt 3 backbones: ResNet-34, ResNet-50, and MobileNetV2 in both Faster RCNN~\cite{ren2015faster} with FPN ~\cite{faster-fpn} and RetinaNet~\cite{lin2017focal}.
We use the CitySpace dataset to benchmark SysNoise on Segmentation Task, where we evaluate two architectures (Deeplabv3 and U-Net). As for DeepLabv3, following~\cite{deeplabv32018}, Resnet-50 and Resnet-101 backbones are used. During Training, we use the Pillow package and choose bilinear as an image resize interpolation method to prepare data. Following~\cite{faster-fpn} ,we resize images by keeping the ratio the same as the original image and make the maximum size of the image to be $1333\times800$. Following common practice, all backbones are pre-trained on ImageNet. We train the default model using FP32 format and train the original upsample layers with the nearest-neighbor interpolation. For the models with the ResNet backbone, we train it with the floor mode of its max-pooling layer. All other implementations follow the original settings of the model.

\textbf{Natural Language Processing Task}
For natural language processing tasks, we use pre-trained OPT ~\cite{opt} models which are transformer-based models with 125M to 175B parameters. For different natural language processing tasks, we use different datasets including PIQA~\cite{piqa}, LAMBADA~\cite{paperno-etal-2016-lambada}, HellaSwag~\cite{hellaswag} and WINOGRANDE~\cite{winogrande}.
Compared with computer vision, natural language tasks have less noise during pre-processing and post-processing progress. For simplicity, we use model inference noise, or data precision noise to measure SysNoise in these tasks.

To benchmark the robustness against SysNoise, we train deep neural networks with one fixed setting, also commonly used in the PyTorch framework, and evaluate the task performance under other settings depending on the different types of SysNoise (~\autoref{benchmark_sysnoise}).

\subsection{Experimental Results}
\label{sec_results}

\textbf{Impact from single SysNoise.}
Our evaluation is summarized in \autoref{tab_cls} for ImageNet classification, \autoref{tab_det} for COCO detection, \autoref{tab_seg} for CitySpace segmentation, and \autoref{tab_nlp} for natural language processing. It can be observed that different types of SysNoise cause different levels of performance drop. \emph{For classification}, The color mode and FP16 precision have a subtle impact on the performance of CNNs, while the image decode and resize can have a 0.6-2.3\% accuracy decrease on average. In model inference noise, the ceiling mode has a profound effect, where the accuracy drops by 0.8-2.7\%. \emph{For detection and segmentation tasks}, there are extra types of SysNoise, the interpolation for upsample layer, and the proposal operation for post-processing. 
Notably, these two types cause a considerable performance drop. They cause Faster RCNN with ResNet-50 backbone drop of 1.7 and 2.4 mAP, respectively. 
\emph{For natural language processing tasks}, the impact of data precision has a greater relationship with the datasets.
In addition, we find SysNoise behaves differently at the task level. For example, The resize noise has a relatively larger impact on the detection task than the classification task. While the decoder noise nearly has no impact on detection and segmentation tasks but can affect classification models. 

\textbf{Architecture-wise robustness again SysNoise.}
We also observe some relationships between architecture and SysNoise for classification. 
\textit{First, in the same architecture family, a larger model tends to have low accuracy degradation}. For instance, in ResNet and RegNet family the average accuracy decrease by decode noise reduces from 1.6\% to 0.6\% when switching from tiny to large models. The same trend is also found in other noises and tasks.
\textit{Second, the lightweight architecture family is more prone to SysNoise.} Specifically, the MobileNetV2 family shows a larger accuracy decrease than other architecture families.
The largest MobileNetV2 drops 1.65\% accuracy due to different resize methods while the similar-accuracy-level ResNet-50 only drops 0.75\%. 
Furthermore, the MCUNet for STM32F746 with just 320KB memory has the worst robustness among all models, which suffers from an average 4.0\% accuracy drop and a maximum 9.3\% accuracy drop in resizing noise.
\textit{Third, ViTs demonstrate different robustness compared with CNNs.} The Swin Transformers are more robust than CNNs when attacked by decoder noise. Interestingly, both ViTs and Swin Transformers suffer from higher accuracy lost in color mode noise than CNNs. These results demonstrate the extremely high diversity of SysNoise.

\begin{table*}[t]
\centering
\caption{\textbf{ Measuring SysNoise on ImageNet classification benchmark.} We record Top-1 accuracy and the difference, $\Delta$ACC = ACC$_{\mathrm{original}}$-ACC$_{\mathrm{SysNoise}}$. We report both mean and max $\Delta$ACC for decode and resize. \emph{The lower} $\Delta$ACC \emph{the better.}}
\adjustbox{max width=\textwidth}{
\begin{tabular}{l c c c c c c c ccccc}
\toprule
\textbf{Architecture} & \multicolumn{1}{c}{\textbf{Trained}} & \multicolumn{1}{c}{\textbf{Decode}} &   \multicolumn{1}{c}{\textbf{Resize}} & \multicolumn{1}{c}{\textbf{Color Mode}}  &  \multicolumn{2}{c}{\textbf{Precision (FP16/INT8)}} & \multicolumn{1}{c}{\textbf{Ceil Mode}} & \multicolumn{1}{c}{\textbf{Combined}} \\
\cmidrule(l{2pt}r{2pt}){3-5} \cmidrule(l{2pt}r{2pt}){6-8} \cmidrule(l{2pt}r{2pt}){9-9} 
& ACC &  $\Delta$ACC & $\Delta$ACC & $\Delta$ACC & $\Delta$ACC & $\Delta$ACC & $\Delta$ACC & $\Delta$ACC \\
\midrule
MCUNet-293KB & 63.40 & 0.41\mypm{0.42} & 4.02\mypm{9.31} & 0.20  & 0.01 & 0.04  & - & 9.97\\
\midrule
ResNet18x0.25 & 48.96  & 1.98\mypm{2.12}  & 2.11\mypm{3.71} & 0.14  & -0.01 & 0.82  & 2.34 & 6.61 \\
ResNet18x0.5 & 61.64 & 1.67\mypm{1.76} & 1.76\mypm{3.25} & 0.19  & -0.01 & 0.15  & 2.72 & 6.10 \\
ResNet-18 & 69.96 & 1.02\mypm{1.03} & 1.01\mypm{2.05} & 0.13  & 0.00 & 0.20  & 2.40 & 4.97 \\
ResNet-34 & 73.59 & 0.99\mypm{1.00} & 0.77\mypm{1.67} & 0.14   & 0.00 & 0.04  & 0.85 & 4.25 \\
ResNet-50 & 76.39   & 0.98\mypm{0.98} & 0.75\mypm{1.69} & 0.09  & 0.00 & 0.06 & 1.24 & 3.95 \\
ResNet-101 & 78.10  & 0.68\mypm{0.69} & 0.62\mypm{1.47} & 0.24   & 0.01 & 0.69  & 0.75 & 4.50 \\
\midrule
MobileNetV2-0.5 & 64.94 & 1.98\mypm{2.00} & 2.04\mypm{3.14} & 0.18 &  0.01 & 0.57 & - & 5.81 \\
MobileNetV2-0.75 & 70.26 & 1.39\mypm{1.39} & 1.47\mypm{2.56} & 0.16   & 0.01 & 0.72 & - & 5.58 \\
MobileNetV2-1 & 73.12 & 1.39\mypm{1.39} & 1.48\mypm{2.43} & 0.07 & 0.02  & 0.77  & - & 5.03  \\
MobileNetV2-1.4 & 75.84 & 1.01\mypm{1.02} &  1.65\mypm{2.15} & 0.10  & 0.01  & 0.53  & - & 5.04   \\
\midrule
RegNetX-400M & 70.97  & 1.63\mypm{1.63} &  1.42\mypm{2.65} & 0.07 & 0.01 & 0.09  & - & 5.70  \\
RegNetX-800M & 74.04  & 1.12\mypm{1.14} &	0.97\mypm{2.00} & 0.19  & 0.02 & 0.24 & - & 4.38  \\
RegNetX-1.6G & 76.29  & 0.84\mypm{0.85} & 0.79\mypm{1.88} & 0.20  & 0.01 & 0.19  & - & 4.15 \\
RegNetX-3.2G & 77.89   & 0.61\mypm{0.62} & 0.53\mypm{1.42} & 0.20  & 0.00 & 0.24 & - & 3.70 \\
\midrule
EfficientNet-B0 & 76.83 & 0.75\mypm{0.76} &  1.70\mypm{3.79} & 0.15 & 0.03 & 0.19  & - & 4.39 \\
EfficientNet-B1 & 78.13 & 0.57\mypm{0.58} &  1.18\mypm{2.84} & 0.26 & 0.01 & 0.39  & - & 3.26 \\
EfficientNet-B2 & 79.97 & 0.57\mypm{0.58} &  1.13\mypm{2.31} & 0.05 & 0.04 & 0.41  & - & 3.10 \\
EfficientNet-B3 & 82.03 & 0.71\mypm{0.72} &  0.99\mypm{1.74} & 0.16 & 0.05 & 0.38  & - & 2.65 \\
EfficientNet-B4 & 83.43 & 0.29\mypm{0.30} &  0.45\mypm{0.93} & 0.17 & 0.02 & 0.26  & - & 2.32 \\
\midrule
ViT-Tiny & 75.61 & 1.04\mypm{1.04} & 0.99\mypm{1.79} & 0.46 & 0.01 & 0.68 & - & 3.21 \\
ViT-Small & 81.58 & 0.57\mypm{0.58} & 0.37\mypm{1.01} & 0.80 & -0.01 & 0.80 & - & 2.68 \\
Vit-Base & 84.63 & 0.61\mypm{0.62} & 0.43\mypm{0.74} & 0.93 & -0.01 & 1.12 & - & 2.89 \\
\midrule
Swin-Tiny & 81.32 & 0.18\mypm{0.19} & 0.42\mypm{1.76} & 1.21 & 0.00 & 0.76 & - & 4.93 \\
Swin-Small & 83.03 & 0.18\mypm{0.18} & 0.23\mypm{1.33} & 1.00 & 0.00 & 0.45 & - & 3.51 \\
Swin-Base & 83.54 & 0.11\mypm{0.30} & 0.21\mypm{1.27} & 0.97 & -0.01 & 0.55 & - & 3.59 \\
\bottomrule
\end{tabular}
}
\label{tab_cls}
\end{table*}

\begin{table*}[htbp]
\centering
\caption{\textbf{Measuring SysNoise on MS COCO detection.} We record mAP and the difference $\Delta$mAP = mAP$_{\mathrm{original}}$-mAP$_{\mathrm{SysNoise}}$. We report both mean and max $\Delta$mAP for decode and resize. \emph{The lower} $\Delta$mAP \emph{the better}. }
\adjustbox{max width=\textwidth}{
\begin{tabular}{l l c c c c c c c c c}
\toprule
\textbf{Method} & \textbf{Architecture} & \multicolumn{1}{l}{\textbf{Trained}} & \multicolumn{1}{c}{\textbf{Decode}} &   \multicolumn{1}{c}{\textbf{Resize}} & \multicolumn{1}{c}{\textbf{Color Mode}}  & \multicolumn{1}{c}{\textbf{Upsample}} & \multicolumn{1}{c}{\textbf{Precision INT8}} & \multicolumn{1}{c}{\textbf{Ceil Mode}}  & \multicolumn{1}{c}{\textbf{Post-processing}} & \multicolumn{1}{c}{\textbf{Combined}} \\
\cmidrule(l{2pt}r{2pt}){4-6} \cmidrule(l{2pt}r{2pt}){7-9} \cmidrule(l{2pt}r{2pt}){10-10} \cmidrule(l{2pt}r{2pt}){11-11} 
& & mAP &  $\Delta$mAP & $\Delta$mAP & $\Delta$mAP & $\Delta$mAP & $\Delta$mAP & $\Delta$mAP  & $\Delta$mAP & $\Delta$mAP \\
\midrule
& ResNet-34 & 36.76 & 0.02\mypm{0.04} & 0.93\mypm{2.63} & 0.25  & 1.28 & 0.06  & 2.50 &  2.29   & 10.25\\
Faster RCNN & ResNet-50 & 37.36  & 0.02\mypm{0.01} & 1.12\mypm{3.15} & 0.10 & 1.66 & 0.10 & 3.14 & 2.39 & 10.67   \\
& MobileNetV2 & 30.32  & 0.01\mypm{0.01} & 0.38\mypm{1.14} & 0.24  & 0.96  & 0.07 & - & 2.23 & 3.45 \\
\midrule
\multirow{2}{*}{RetinaNet} & ResNet-34 &  35.71  & 0.01\mypm{0.01} & 0.77\mypm{2.20} & 0.29  & 0.35 & 0.10 & 2.72 & 3.44 & 8.21  \\
& ResNet-50&  36.59 & 0.01\mypm{0.02} & 0.99\mypm{2.78} & 0.36 & 0.69 & 0.03 & 3.12 & 3.00 & 8.93  \\
\bottomrule
\end{tabular}
}
\label{tab_det}
\end{table*}

\begin{table*}[htbp]
\centering
\caption{\textbf{Measuring SysNoise on CitySpace segmentation.} We record mIOU and the difference $\Delta$mIOU = mIOU$_{\mathrm{original}}$-mIOU$_{\mathrm{SysNoise}}$. We report both mean and max $\Delta$mIoU for decode and resize. \emph{The lower} $\Delta$mIOU \emph{the better}. }
\adjustbox{max width=\textwidth}{
\begin{tabular}{l l c c c c c c c c}
\toprule
\textbf{Method} & \textbf{Architecture} & \multicolumn{1}{l}{\textbf{Trained}} & \multicolumn{1}{c}{\textbf{Decode}} &   \multicolumn{1}{c}{\textbf{Resize}} & \multicolumn{1}{c}{\textbf{Color Mode}}  & \multicolumn{1}{c}{\textbf{Upsample}} & \multicolumn{1}{c}{\textbf{Precision INT8}} & \multicolumn{1}{c}{\textbf{Ceil Mode}} & \multicolumn{1}{c}{\textbf{Combined}} \\
\cmidrule(l{2pt}r{2pt}){4-6} \cmidrule(l{2pt}r{2pt}){7-9} \cmidrule(l{2pt}r{2pt}){10-10} 
& & mIoU &  $\Delta$mIoU & $\Delta$mIoU & $\Delta$mIoU & $\Delta$mIoU & $\Delta$mIoU & $\Delta$mIoU & $\Delta$mIoU \\
\midrule
\multirow{2}{*}{DeepLabV3} & ResNet-50 & 78.05 & 0.001\mypm{0.001} & 0.02\mypm{0.04} & 0.02  & 3.06 & 0.01  & 4.02 & 4.51 \\
& ResNet-101 & 79.88 & 0.001\mypm{0.001} & 0.01\mypm{0.02} & 0.02  & 3.85 & 0.01  & 4.65 & 5.11 \\
\midrule
\multirow{1}{*}{U-Net} & - & 61.98 & 0.003\mypm{0.005} & 0.04\mypm{0.06} & 0.04  & 2.74 & 0.02  & - & 2.85 \\
\bottomrule
\end{tabular}
}
\label{tab_seg}
\end{table*}

\begin{table}[htbp]
\centering
\caption{\textbf{Measuring SysNoise on Multiple NLP Datasets.} We record ACC on FP32 data precision and the difference $\Delta$ACC = ACC$_{\mathrm{original}}$-ACC$_{\mathrm{SysNoise}}$ on FP16 and INT8 data precision. \emph{The lower} $\Delta$ACC \emph{the better}. }
\adjustbox{max width=\textwidth/2}{
\begin{tabular}{l c c c c}
\toprule
\multicolumn{1}{l}{\textbf{Architecture}} & \multicolumn{1}{c}{\textbf{PIQA}} & \multicolumn{1}{c}{\textbf{LAMBADA}} & \multicolumn{1}{c}{\textbf{HellaSwag}}  & \multicolumn{1}{c}{\textbf{WINOGRANDE}} \\
\cmidrule(l{2pt}r{2pt}){2-5} 
\multicolumn{1}{c}{} & \multicolumn{4}{c}{FP32(ACC)/FP16($\Delta$ACC)/INT8($\Delta$ACC)} \\
\midrule
OPT-125M & 63.00/0.05/-0.06 & 37.90/0.04/0.37 & 29.20/0.01/0.15 & 50.28/0.00/-0.31 \\
OPT-350M & 64.36/-0.11/-0.33 & 45.16/0.00/-0.10 & 32.04/0.00/0.05 & 52.33/-0.08/0.24 \\
OPT-1.3B & 71.71/-0.05/0.16 & 58.06/0.07/0.19 & 41.45/-0.03/0.08 & 59.67/0.00/-0.24 \\
OPT-2.7B & 73.78/0.06/0.16 & 63.65/0.09/0.02 & 45.85/-0.01/0.01 & 61.01/0.00/0.00 \\
OPT-6.7B & 76.06/0.16/0.22 & 67.61/0.06/0.33 & 50.46/0.01/0.03 & 65.04/0.00/0.48 \\
OPT-13B & 75.90/0.11/0.06 & 68.72/0.07/0.29 & 52.44/-0.02/0.01 & 65.11/0.00/0.39 \\
OPT-30B & 77.69/0.11/0.16 & 71.47/0.02/0.00 & 54.29/-0.01/0.01 & 68.19/0.02/0.23 \\
\bottomrule
\end{tabular}
}
\label{tab_nlp}
\end{table}

\textbf{Impact from multiple SysNoise.}
The single noise type may only have limited impacts on task performance. However, it is likely that SysNoise will happen in multiple stages during inference, and will have a combined effect with multiple noises to bring further influences on the accuracy.

\begin{figure}[htbp]
    \centering
    \subfigure[ResNet-50 ImageNet]{\includegraphics[width = 0.48\linewidth]{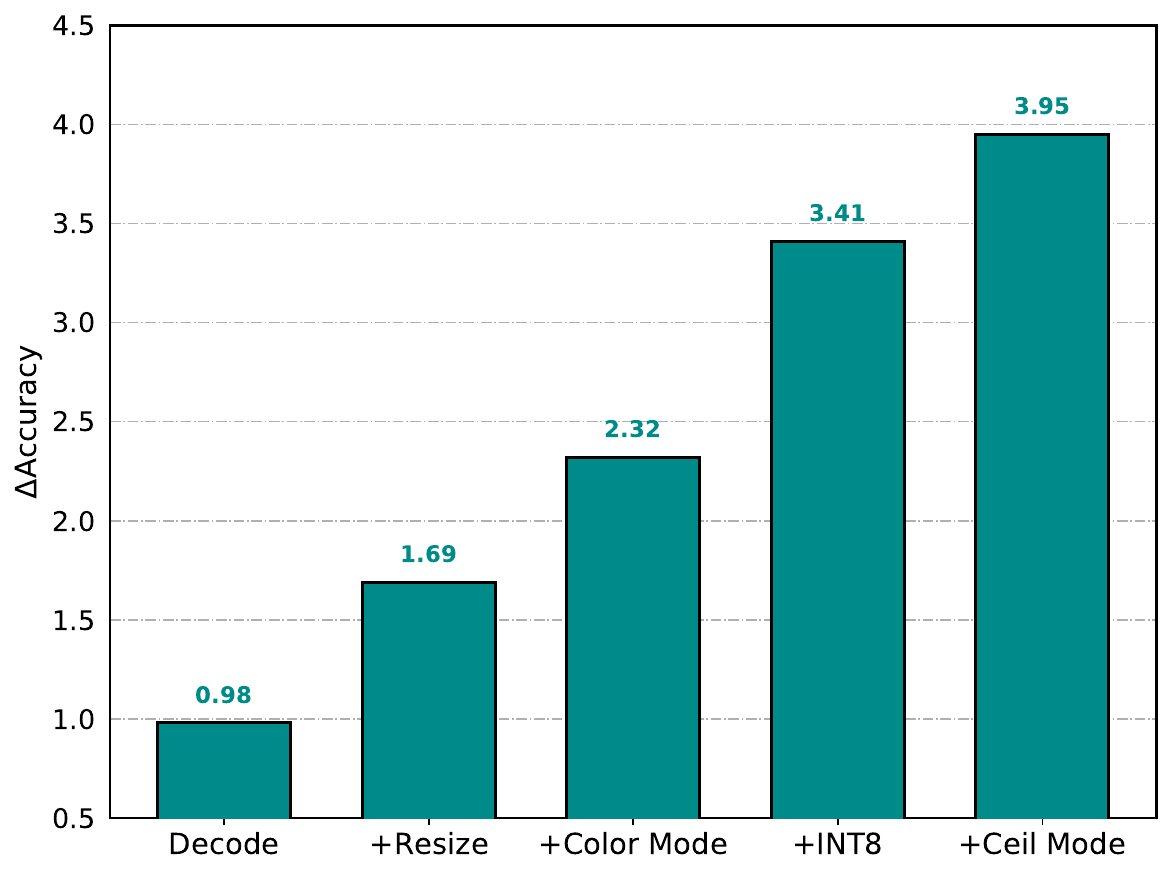}} 
    \subfigure[Faster-RCNN ResNet-50 COCO]{\includegraphics[width = 0.5\linewidth]{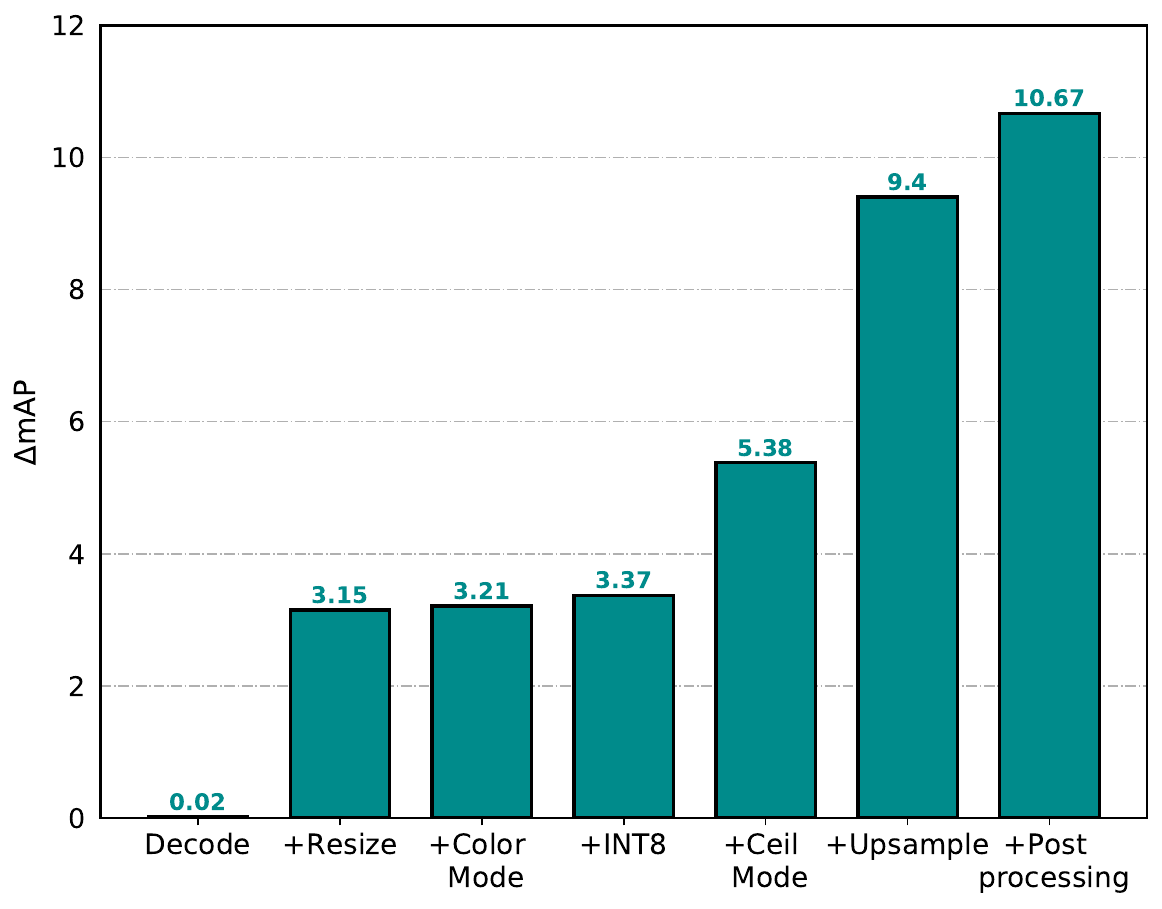}}
    
    \caption{\textbf{Illustration of the worst-case study by combining multiple SysNoise types step by step}.}
    \label{fig_wc}
\end{figure}

We show how combined SysNoise affects a single model step by step in \autoref{fig_wc}. For example, on ResNet-50, we select the most influential SysNoise type and gradually add them to impose noise coherently. As shown in \autoref{fig_wc}, we show that some SysNoise is lessened while others are strengthened when combined together. For instance, adding resize to ResNet-50 incurs 0.71\% extra accuracy loss which is even lower compared to the average 0.75\% accuracy loss in \autoref{tab_cls}. On the contrary, the INT8 quantization increases its damage from 0.06\% to 1.09\%. This reveals two discoveries. First, different types of pre-processing noise can overlap with each other. Second, model inference noise might be magnified with other noises. We will provide more in-depth future studies. Interestingly, we show that the impact from SysNoise can be magnified especially in detection tasks whose model has ceil mode and upsample noise together. We deduce that this may be because they are both noises about the relative position and value of the model's feature map and the superposition of these two noises can cause effects beyond their own noise.

In \autoref{tab_cls}, \autoref{tab_det} and \autoref{tab_seg}, we show how combined SysNoise affects different models on different model architecture.
As shown in \autoref{tab_cls} and \autoref{tab_det}, adding all SysNoise to ResNet-50 together can damage 3.95\% accuracy for classification and 10.67\% mAP for detection, which equals degenerating a ResNet-50 lower than ResNet-34. Adding all SysNoise to EfficientNet-B4 makes it lower than the B3 variant. According to the original paper~\cite{tan2020efficientnet}, B4 consumes 2.3$\times$ more FLOPs than B3 and 1.6$\times$ higher parameters, yet only 1.4\% accuracy improvement. However, SysNoise can easily make the architecture improvement useless, with up to 2.3\% accuracy degradation to EfficientNet-B4.

\subsection{Interpreting SysNoise}

\label{sec_interp}

\begin{figure*}[htbp]
    \centering
    \includegraphics[width=0.95\linewidth]{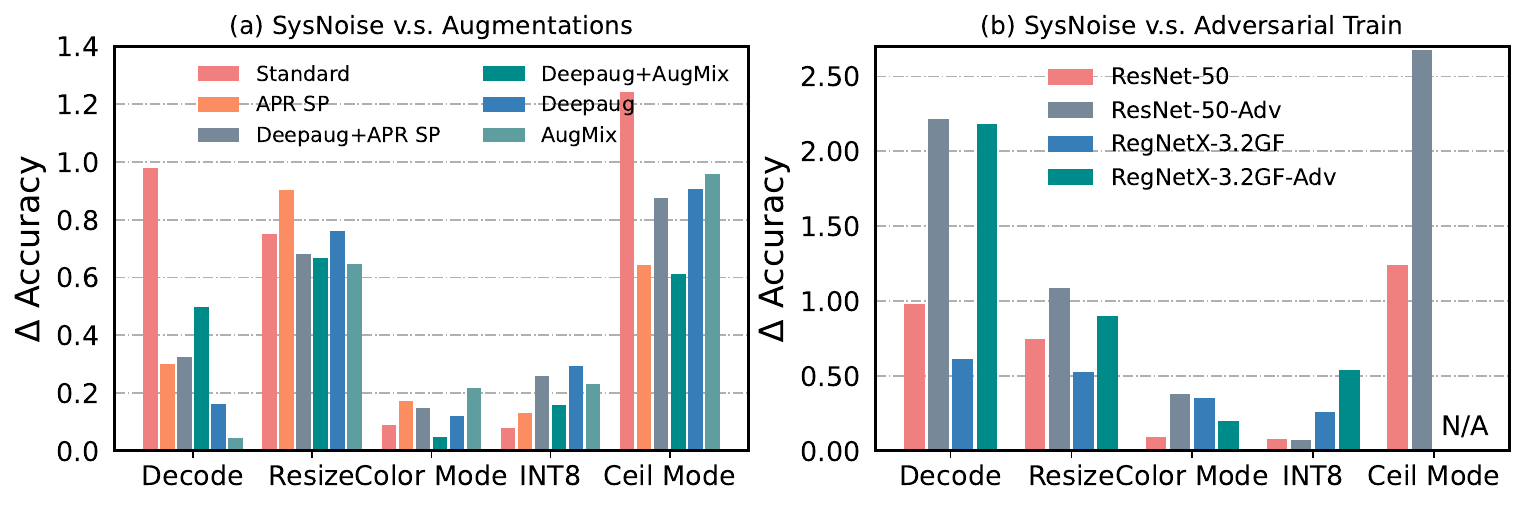}
    \caption{\textbf{Illustration of data augmentations and adversarial training for SysNoise on ImageNet}. }
    \label{fig_aug_adv}
\end{figure*}

\begin{figure*}[htbp]
\adjustbox{margin=1em,width=\textwidth,set height=0.4cm,set depth=0.1cm,center}{Original Image\ \ \ \ \ \ \ \ \ \ \ \ \ Decode\ \ \ \ \ \ \ \ \ \ \ \ \ \ \ \ \ \ Resize\ \ \ \ \ \ \ \ \ \ \ \ \ \ \ \ \ Color Mode\ \ \ \ \ \ \ \ \ \ \ \ \ \ \ \ \ INT8\ \ \ \ \ \ \ \ \ \ \ \ \ \ \ \ \ \ \ \ Ceil Mode\ \ }
\subfigure{
\includegraphics[width=.15\linewidth]{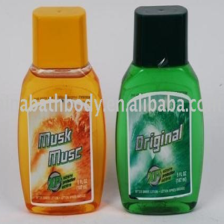}
}
\subfigure{
\includegraphics[width=.15\linewidth]{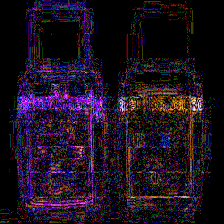}
}
\subfigure{
\includegraphics[width=.15\linewidth]{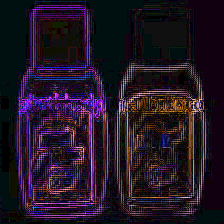}
}
\subfigure{
\includegraphics[width=.15\linewidth]{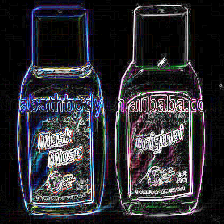}
}
\subfigure{
\includegraphics[width=.15\linewidth]{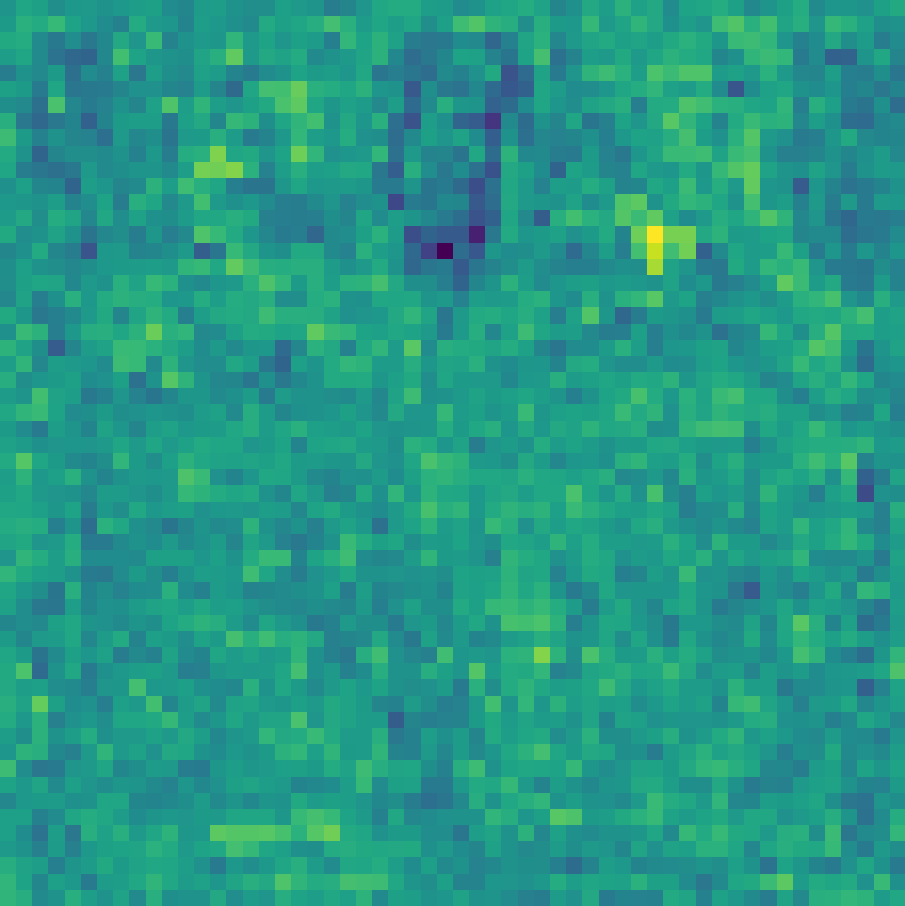}
}
\subfigure{
\includegraphics[width=.15\linewidth ]{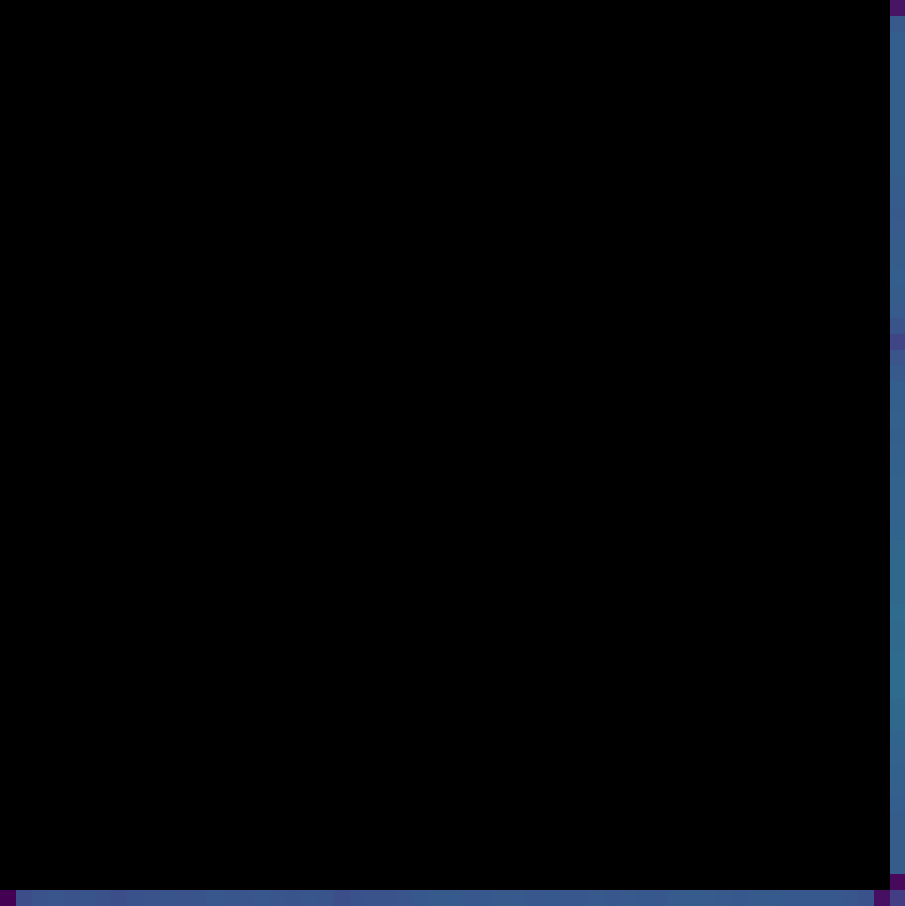}
}
\\
\vskip -15pt
\subfigure{
\includegraphics[width=.15\linewidth]{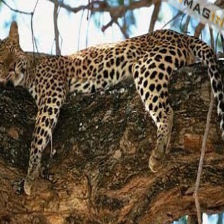}
}
\subfigure{
\includegraphics[width=.15\linewidth]{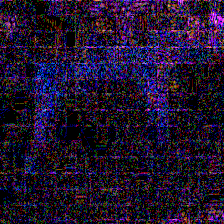}
}
\subfigure{
\includegraphics[width=.15\linewidth]{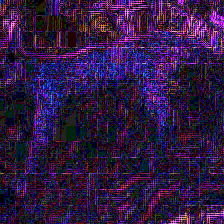}
}
\subfigure{
\includegraphics[width=.15\linewidth]{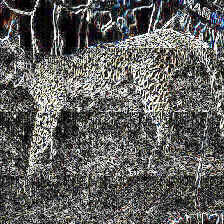}
}
\subfigure{
\includegraphics[width=.15\linewidth]{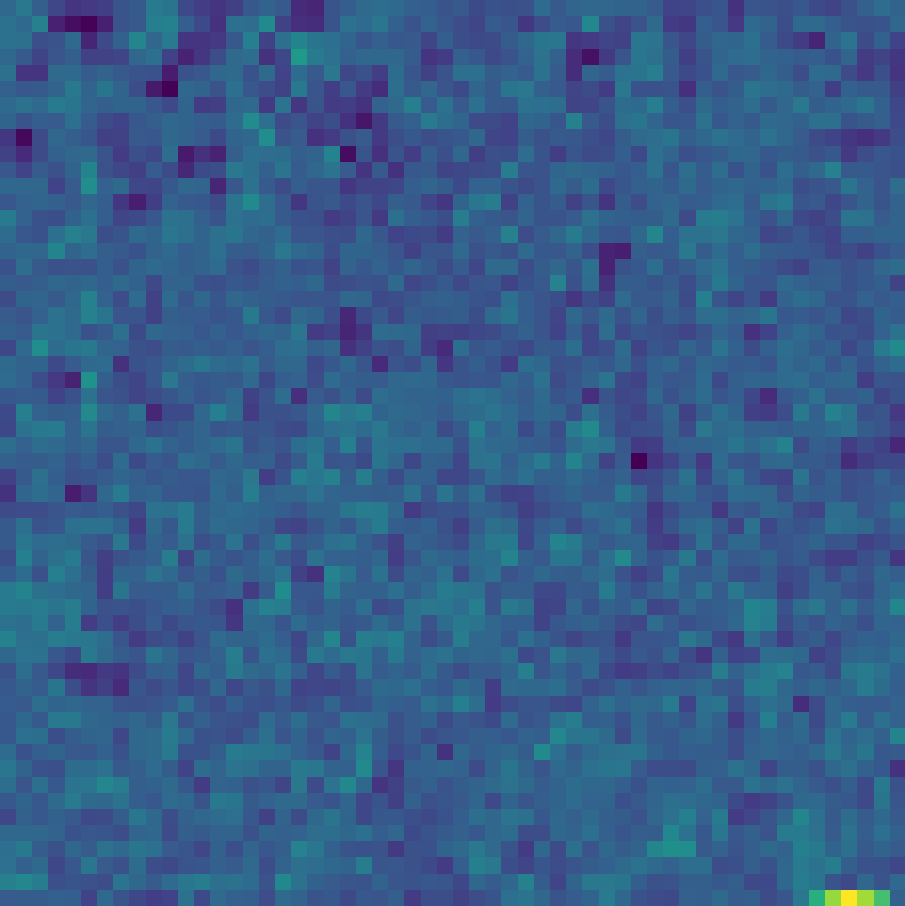}
}
\subfigure{
\includegraphics[width=.15\linewidth ]{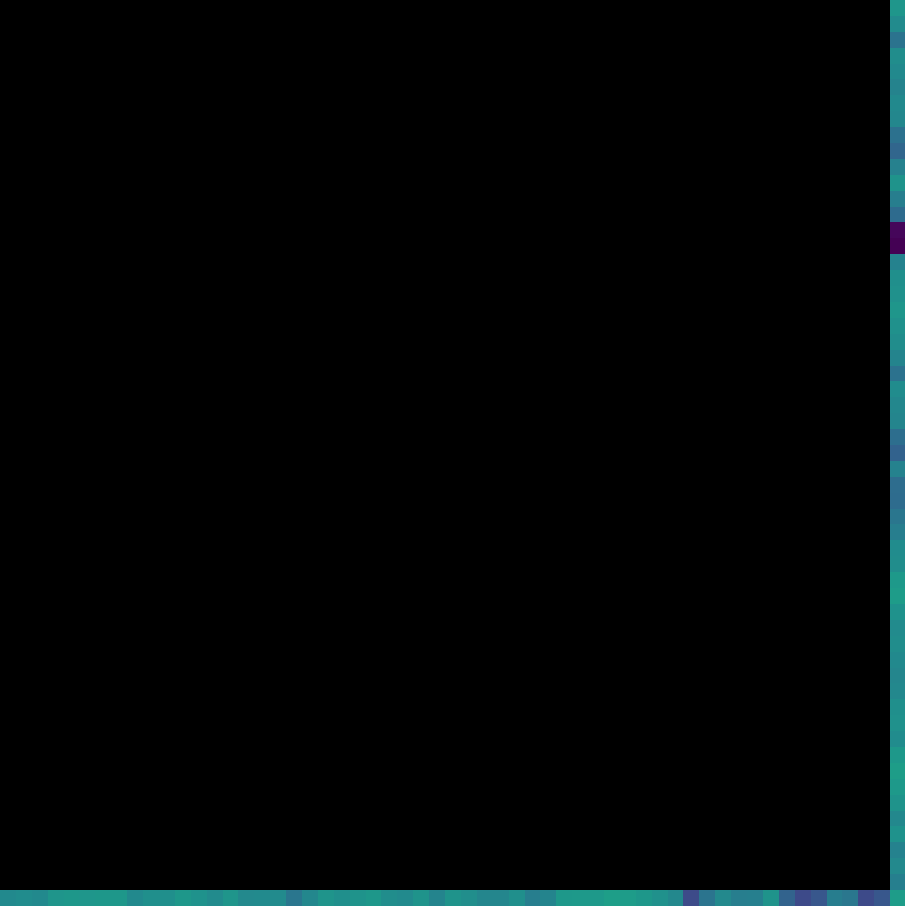}
}
\\
\vskip -15pt
\subfigure{
\includegraphics[width=.15\linewidth]{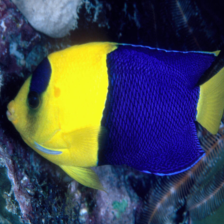}
}
\subfigure{
\includegraphics[width=.15\linewidth]{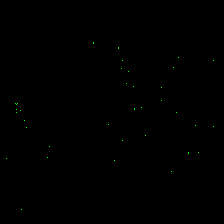}
}
\subfigure{
\includegraphics[width=.15\linewidth]{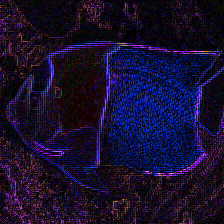}
}
\subfigure{
\includegraphics[width=.15\linewidth]{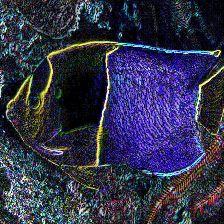}
}
\subfigure{
\includegraphics[width=.15\linewidth]{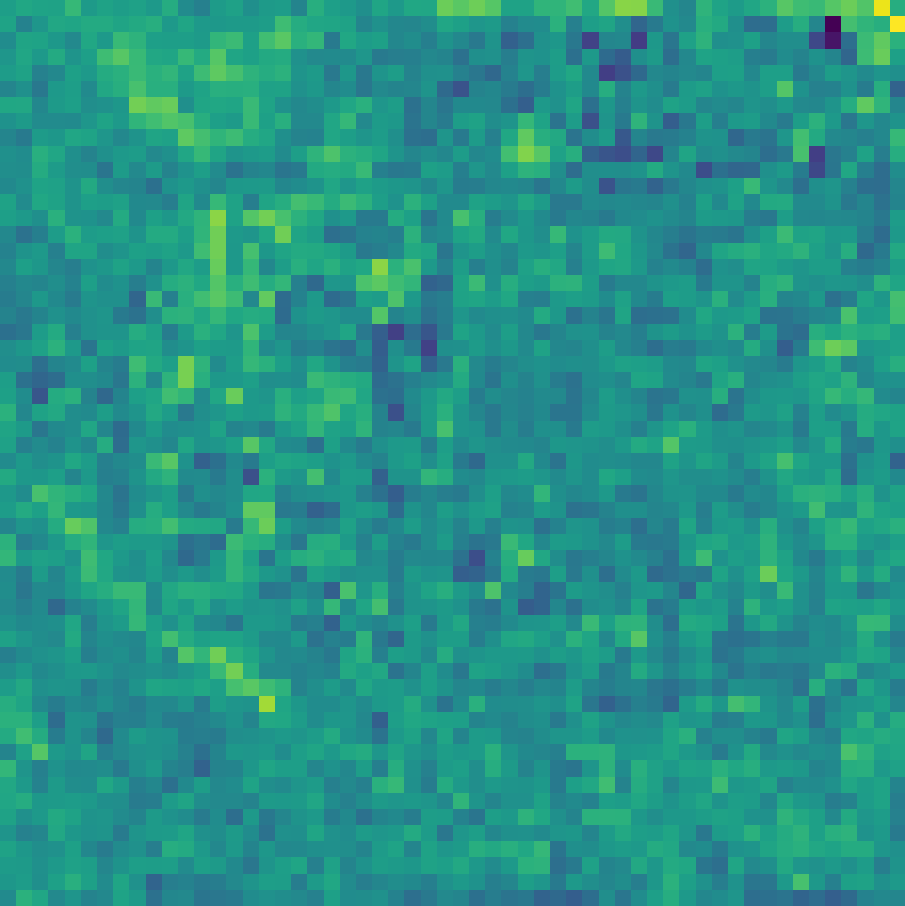}
}
\subfigure{
\includegraphics[width=.15\linewidth ]{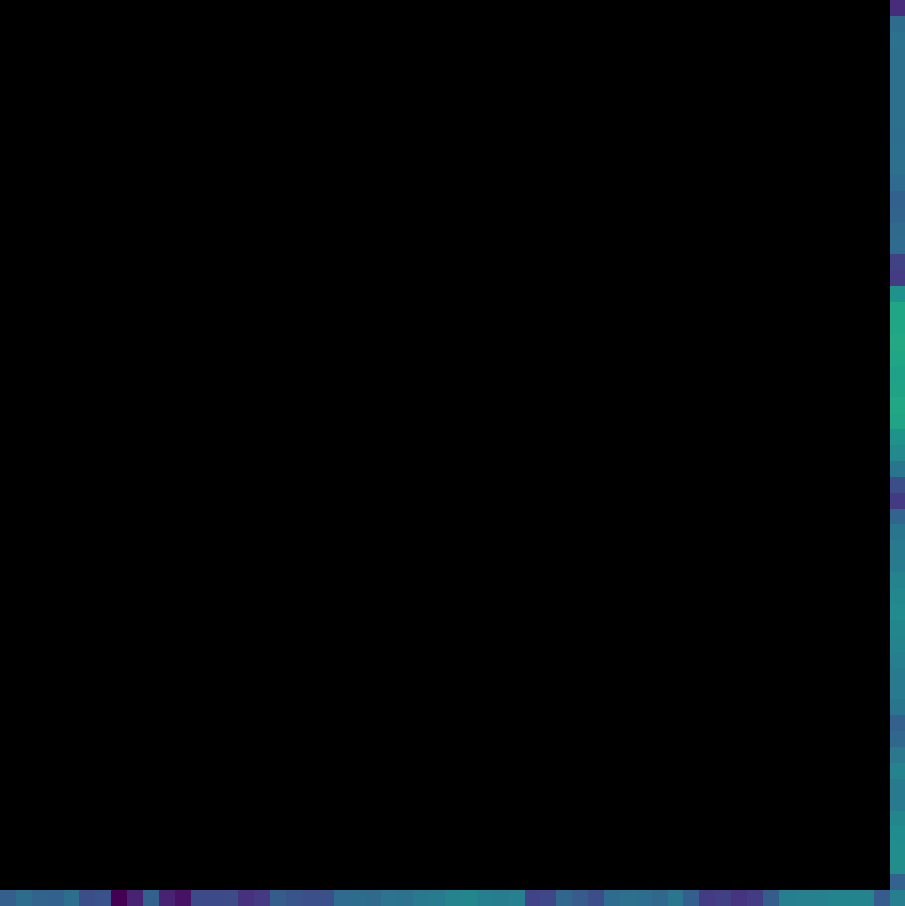}
}
\\
\vskip -15pt
\subfigure{
\includegraphics[width=.15\linewidth]{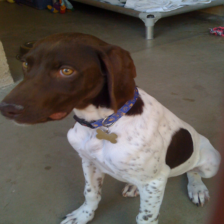}
}
\subfigure{
\includegraphics[width=.15\linewidth]{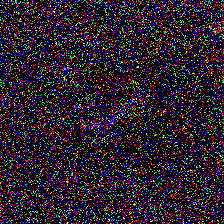}
}
\subfigure{
\includegraphics[width=.15\linewidth]{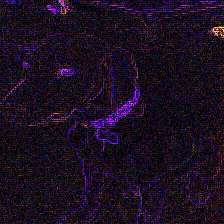}
}
\subfigure{
\includegraphics[width=.15\linewidth]{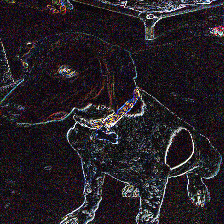}
}
\subfigure{
\includegraphics[width=.15\linewidth]{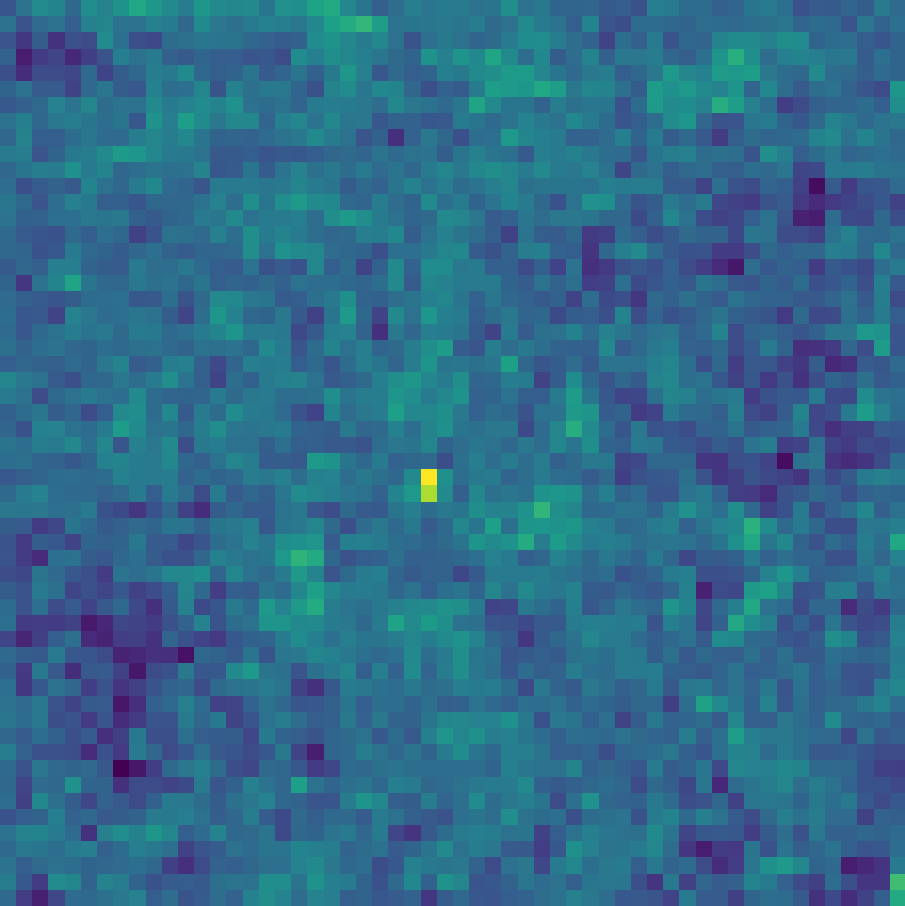}
}
\subfigure{
\includegraphics[width=.15\linewidth ]{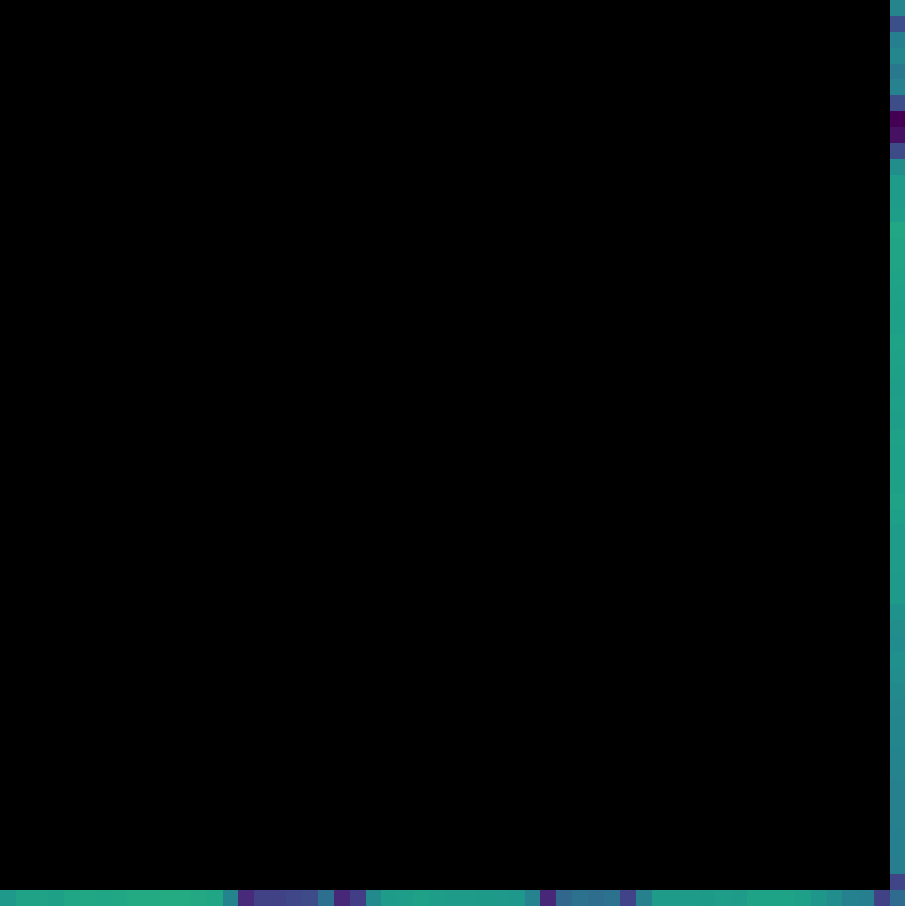}
}
\caption{\textbf{Visualization of SysNoise. To make the noise more perceptible, we scale it to [0, 255]}.}
\label{fig:vis}
\end{figure*}

\textbf{Does data augmentation improve model robustness against SysNoise?} 
Studies have shown that data augmentation techniques can be used to improve model robustness against natural noises~\cite{hendrycks2020augmix,hendrycks2021nae}. We, therefore, employ these augmentation methods to see whether they could also improve the robustness against SysNoise. In particular, we train a ResNet-50 model using standard augmentation~\cite{DBLP:journals/corr/HeZRS15}, APR-SP~\cite{chen2021amplitude}, Deepaug~\cite{hendrycks2020many}, AugMix~\cite{hendrycks2020augmix}, and Deepaug combined with the other two methods (denoted ``Deepaug+APR-SP'' and ``Deepaug+AugMix''). As shown in \autoref{fig_aug_adv}, we could reach several observations as follows (1) \emph{there exist no single data augmentation methods that could universally achieve positive effects on all of the five different SysNoise types}; (2) specifically, data augmentations could improve model robustness against image decoder, the ceiling mode of the max-pooling layers (lower $\Delta$ACC). However, they fail to generalize for data precision and image resize (higher $\Delta$ACC). These indicate that SysNoise is highly diverse and inherently different from natural noises.

\textbf{Does adversarial training improve model robustness against SysNoise?}
Besides natural noises, another axis to analyze SysNoise is to examine whether adversarially-robust models could be also effective against SysNoise. Here, we use adversarial training (the most effective method to defend adversarial noises)~\cite{PGD}, and adversarially train ResNet-50 and RegNetX-3.2GF models with $\ell_{\infty}$-PGD attacks~\cite{PGD} using the standard setting~\cite{tang2021robustart,croce2020robustbench}. The results are summarized in \autoref{fig_aug_adv}, from which we can tell that \emph{adversarial training has limited effect on improving model robustness against SysNoise} (significantly higher $\Delta$ACC on 80\% SysNoise types). In some cases, like image decode and resize, adversarial training even significantly damages the model performance on SysNoise (significantly high $\Delta$ACC). Together with the data augmentation analysis, we show that SysNoise differs from both natural noises and adversarial noise, and the effectiveness of defenses that are designed for adversarial and natural noises is limited for SysNoise. We hope all these observations could inspire more in-depth future studies on building robust models against SysNoise.

\textbf{Does test-time adaptation improve model robustness against SysNoise?}
Different from data augmentation and adversarial training that try to solve issues during the training process, test-time adaptation tries to solve data shifts problem during the testing process. A fully test-time adaptation method called TENT \cite{tent} was proposed which is taken effect by minimizing the entropy of model predictions during model inference. Experiments were carried out to find whether test-time adaptation improves model robustness against SysNoise. And results are shown in \autoref{tab_tent}, from which we can tell that TENT harms the model robustness against SysNoise, except the ViT model zoo on color mode noise. It may be because the data shifts caused by SysNoise are so small compared with that caused by other corruption mentioned in this paper that test-time adaptation harms the performance of the model.

\begin{table}[t]
\centering
\caption{\textbf{ Measuring SysNoise on models with/without TENT.} We record Top-1 accuracy and the difference, $\Delta$ACC = ACC$_{\mathrm{original}}$-ACC$_{\mathrm{SysNoise}}$. We report both mean and max $\Delta$ACC for decode and resize. \emph{The lower} $\Delta$ACC \emph{the better.}}
\adjustbox{max width= \textwidth/2}{
\begin{tabular}{l c c c c c c c ccccc}
\toprule
\textbf{Architecture} & \multicolumn{1}{c}{\textbf{Trained}} & \multicolumn{1}{c}{\textbf{Decode}} &   \multicolumn{1}{c}{\textbf{Resize}} & \multicolumn{1}{c}{\textbf{Color Mode}}\\  
& ACC &  $\Delta$ACC & $\Delta$ACC & $\Delta$ACC  \\
\midrule
MCUNet-293KB(w/o TENT) & 63.40 & 0.41\mypm{0.42} & 4.02\mypm{9.31} & 0.20 \\
MCUNet-293KB(w/ TENT) & 63.40 & 5.03\mypm{7.25} & 4.76\mypm{6.22}  & 0.95\\
\midrule
ResNet-18(w/o TENT) & 69.96 & 1.02\mypm{1.03} & 1.01\mypm{2.05} & 0.13 \\
ResNet-18(w/ TENT) & 69.96 & 4.01\mypm{4.22} & 3.33\mypm{4.06} & 2.16 \\
ResNet-34(w/o TENT) & 73.59 & 0.99\mypm{1.00} & 0.77\mypm{1.67} & 0.14  \\
ResNet-34(w/ TENT) & 73.59 & 4.02\mypm{4.39} & 2.99\mypm{3.51} & 1.89  \\
ResNet-50(w/o TENT) & 76.39   & 0.98\mypm{0.98} & 0.75\mypm{1.69} & 0.09 \\
ResNet-50(w/ TENT) & 76.39   & 4.81\mypm{5.44} & 4.67\mypm{5.41} & 3.78 \\
\midrule
MobileNetV2-0.5(w/o TENT) & 64.94 & 1.98\mypm{2.00} & 2.04\mypm{3.14} & 0.18  \\
MobileNetV2-0.5(w/ TENT) & 64.94 & 7.30\mypm{8.24} & 3.67\mypm{4.76} & 1.04  \\
MobileNetV2-1(w/o TENT) & 73.12 & 1.39\mypm{1.39} & 1.48\mypm{2.43} & 0.07  \\
MobileNetV2-1(w/ TENT) & 73.12 & 5.58\mypm{6.09} & 2.26\mypm{3.07} & 0.77  \\
\midrule
ViT-Tiny(w/o TENT) & 75.61 & 1.04\mypm{1.04} & 0.99\mypm{1.79} & 0.46 \\
ViT-Tiny(w/ TENT) & 75.61 & 1.38\mypm{1.57} & 1.16\mypm{1.81} & 0.28 \\
Vit-Base(w/o TENT) & 84.63 & 0.61\mypm{0.62} & 0.43\mypm{0.74} & 0.93 \\
Vit-Base(w/ TENT) & 84.63 & 1.71\mypm{1.91} & 1.07\mypm{1.37} & 0.90 \\
\midrule
Swin-Tiny(w/o TENT) & 81.32 & 0.18\mypm{0.19} & 0.42\mypm{1.76} & 1.21  \\
Swin-Tiny(w/ TENT) & 81.32 & 7.32\mypm{9.11} & 3.68\mypm{4.95} & 2.28  \\
Swin-Base(w/o TENT) & 83.54 & 0.11\mypm{0.30} & 0.21\mypm{1.27} & 0.97 \\
Swin-Base(w/ TENT) & 83.54 & 5.98\mypm{6.57} & 3.43\mypm{4.47} & 2.68 \\
\bottomrule
\end{tabular}
}
\label{tab_tent}
\end{table}


\textbf{Potential methods to improve robustness against SysNoise.}
To solve SysNoise on the decoder and resize, a natural way is to make the model "see" all kinds of decoders and resize methods during the training process. Based on this principle, we introduce \textit{mix training} method to enhance the model's robustness on system noise. The main process of mix training is to select the decoder or resize method randomly instead of just using one kind of method during the whole process of training. The pseudocode of our algorithm is shown in
\autoref{algo1}.

To test the effect of mix training, we set up the following experiment. We use ResNet50 as the base model of this experiment.
To comprehensively demonstrate the training effect, we train single decoding and resize as well as our mix training models. 
We set the default decoder as Pillow and the default resize method as Pillow bilinear when conducting ablation studies on resizing method or decoder, respectively. Then we use top-1 accuracy as well as their mean and standard deviation as assessments.
The results of this experiment are shown in \autoref{tab:mix_decoder} and \autoref{tab:mix_resize}. From these tables, we can conclude that: (1) The model has a better performance (usually the best) when we train and test using the same decoder and resize method. (2) Mix training can improve the robustness of a model on system noise greatly without hurting the clean accuracy. The $Std.$ using mix training drop from $0.36$ to $0.0653$ on decoder experiment, and drop from $0.463$ to $0.270$ on resize experiment. Meanwhile, it can maintain the model's accuracy at about 76\%. In a contrast, the same ResNet50 model using $L_{\infty}-Robust$ adversarial training drops the $Std.$ from $1.07$ to $0.420$ by paying a $19.2\%$ drop of clean accuracy.

\begin{table*}[h]
\caption{Mix training on resize method.}
\label{tab:mix_resize}
\adjustbox{max width=\textwidth}{
\begin{tabular}{l|llllll|ll}
\toprule
\diagbox{Train}{Test} & Pillow-bilinear & Pillow-nearest & Pillow-cubic & OpenCV-nearest & OpenCV-bilinear & OpenCV-cubic & Mean & Std. \\
\midrule
Pillow-bilinear  & 76.572 & 72.168  & 76.512  & 72.090          & 75.346      & 74.072    & 74.460 & 2.02E+00 \\
Pillow-nearest               & 74.872       & 75.988      & 75.548    & 75.970          & 76.002          & 76.056       & 75.739	& 4.63E-01 \\
Pillow-cubic                 & 76.312       & 72.828      & 76.596    & 72.876         & 75.810           & 74.666       & 74.848	& 1.68E+00 \\
OpenCV-nearest            & 74.818       & 76.298      & 75.474    & 76.092         & 76.082          & 76.192       & 75.826	& 5.71E-01\\
OpenCV-bilinear           & 75.840        & 75.268      & 76.446    & 75.248         & 76.682          & 76.436       & 75.987 & 6.29E-01 \\
OpenCV-cubic              & 76.194       & 72.812      & 76.510     & 72.940          & 75.736          & 74.818       & 74.835 & 1.62E+00 \\
mix                       & 76.154       & 75.876      & 76.344    & 75.786         & 76.444          & 76.330        & \textbf{76.156} & \textbf{2.70E-01} \\
\bottomrule
\end{tabular}
}
\end{table*}

\begin{algorithm}[h]
\caption{Mixed training for improving robustness on systematic noise.}
\label{algo1}
\KwIn{Resize set $\mathbb{RS}$; Decoder set $\mathbb{D}$; Model to train}
Set Pillow-bilinear as default Resize; \\
Set Pillow as default Decoder; \\
\For{all $j=1,2,\dots,T$-iteration in training}
{
    \If{use mix-decoder strategy}{
        Randomly sample a Decoder from $\mathbb{D}$;
    }
    \If{use mix-resize strategy}{
        Randomly sample a Resize function from $\mathbb{RS}$;
    }
    Load the images from the file system according to the Decoder type and Resize type; \\
    Model Optimization.
}    
\KwRet{An optimized robust model for systematical noise.}

\end{algorithm}

\begin{table}[h]
\centering
\caption{Mix training on the decoder.}
\label{tab:mix_decoder}
\adjustbox{max width=0.48\textwidth}{
\begin{tabular}{l|ccc|cc}
\toprule
\diagbox{Train}{Test} & Pillow            & OpenCV          & FFmpeg          & Mean & Std.               \\
\midrule
Pillow  & 76.430  & 76.426 & 75.310 & 76.055 & 6.45E-01          \\
OpenCV  & 76.510   & 76.510  & 75.368   & 76.126 & 6.56E-01   \\
FFmpeg & 75.730  & 75.664 & 76.318 & 75.904 & 3.60E-01  \\
mix  & \textbf{76.53} & \textbf{76.524}  & \textbf{76.414} & \textbf{76.489} & \textbf{6.53E-02}\\
\bottomrule
\end{tabular}
}
\end{table}

\textbf{Visualization.} Here, we visualize the SysNoise by showing the \emph{difference in pixels}. In specific, we calculate the differences between the clean image (or feature) and corrupted ones using SysNoise. As shown in \autoref{fig:vis}, we can draw several interesting observations as follows. For the decode noise, it seems to be irregular (totally random or centered around the edge). As for resize and color mode noise, we observe that the differences often appear in edges or corners of an object (\ie, shape). Specifically, Resize noise tends to mismatch in the red channel while color mode noise mismatches in all 3 channels. For Ceil Mode noise, it injects two bands of noises at the bottom right of the image. There is no obvious pattern for the INT8 noise.

\section{Conclusion}
This paper introduces SysNoise, a harmful noise that frequently happens when the source training system switches to a disparate target system in deployments. We first identify and classify SysNoise based on the inference stage, and thereafter build a holistic benchmark and framework to quantitatively measure the impact of SysNoise on image classification, object detection, segmentation, and natural language processing tasks. Our large-scale experiments revealed that SysNoise is highly-influential and will cause model performance degeneration; additionally, common mitigations like data augmentation and adversarial training show limited effects on SysNoise.  

In the future, we will evaluate SysNoise on the real-world systems, and will continuously develop the benchmark to include more tasks. Our findings open a new research topic and we hope it will raise research attention to the performance and robustness of deep learning deployment systems. 


\section*{Acknowledgement}
This work was supported by the National Key Research and Development Plan of China (2021ZD0110601), the National Natural Science Foundation of China (62022009 and 62206009), and the State Key Laboratory of Software Development Environment.

\clearpage
\nocite{langley00}

\bibliography{example_paper}
\bibliographystyle{mlsys2023}


\clearpage

\appendix
\section{Mathematical Difference of SysNoise}
\label{appendix_math}
In this section, we try to use equations to describe how different processing, operations are formulated. Note that our explanation might not be exactly the same with third-party implementations, as there are always some hyper-parameters to determine. Our goal is to provide an intuition rather than a strict comparison. 

\textbf{\emph{Image Decode.}} In the decoding process, the inverse discrete cosine transform (iDCT) occupies the majority of the computation. Given a transformed matrix $\hat{\rmX}$ with shape $N\times N$ (excluding channels), the original image $\rmX$ at coordinates $(m, n)$ can be given by
\begin{equation}
\begin{split}
  f[m,n]=\sum_{k=0}^{N-1}\sum_{l=0}^{N-1}&\alpha(k)\alpha(l)F(k,l)\\
  &cos[\frac{(2m+1){\pi}k}{2N}]cos[\frac{(2n+1){\pi}l}{2N}] \\
\end{split}
\end{equation}
where,
\begin{equation}
  \alpha(k)= \begin{cases}
  \sqrt{\frac{1}{N}} \, \text{  if  } \, k = 0 \\
  \sqrt{\frac{2}{N}} \, \text{  if  } \, k \neq 0
  \end{cases}.
\end{equation}
The iDCT costs a lot of operations and some implementations choose to utilize Fast DCT and Fast iDCT~\cite{chen1977fast} where the computation is sped up by matrix decomposition. Due to its complexity, we do not display the equations here. Note that the de-quantization in decode will also bring different values, which will be introduced in the data precision section. 

\textbf{\emph{Resize Interpolation.}} Formally, considered an image $\rmX$ to be resized where a pixel in some position needs to be predicted and yet its neighbors are already known or predicted. Different interpolation algorithms rely on different functions to determine the unknown pixel. (1) Nearest interpolation, this method simply copy the nearest neighbor's pixel value, \ie, the neighbor with the lowest Euclidean distance, given by $\rmX[\arg\min_{x, y}((x-x^{\prime})^2 +(y-y^{\prime})^2 )]$. Here, the $x, y$ is the coordinates of the known neighbor and $x^{\prime}, y^{\prime}$ is the coordinates of the pixel that needs to be determined. (2) Bilinear interpolation, determines the pixel by linearly calculating the ratio of distance. Assume we have four spatially-close coordinates:  $Q_{11} = (x1, y1)$, $Q_{12} = (x1, y2)$, $Q_{21} = (x2, y1)$, and $Q_{22} = (x2, y2)$. Their values are already know, for example $f(Q_{11})$. The formulation of bilinear interpolation is given by:
\begin{equation}
    f(x, y) = \frac{y_2 - y}{y_2 - y_1}f(x, y_1) + \frac{y - y_1}{y_2 - y_1}f(x, y_2),
\end{equation}
where,
\begin{align}
    f(x, y_1) &= \frac{x_2 - x}{x_2 - x_1}f(Q_{11}) + \frac{x - x_1}{x_2 - x_1}f(Q_{21}), \notag \\
    f(x, y_2) &= \frac{x_2 - x}{x_2 - x_1}f(Q_{12}) + \frac{x - x_1}{x_2 - x_1}f(Q_{22}). \notag
\end{align}
(3) Bicubic interpolation, in contrast to the bilinear interpolation which only takes 4 pixels ($2\times 2$), the bicubic interpolation takes 16 pixels ($4\times 4$). The algorithm tries to use existing known pixel values to fit a binary cubic function
\begin{equation}
    f(x, y) = \sum_{i=0}^3 \sum_{j=0}^3 a_{ij}x^iy^j
\end{equation}
To find the total 16 coefficients $a_{ij}, ij\in\{0,1,2,3\}$, we need to solve a system of linear equations $A\alpha=x$. Due to the complexity of this algorithm, we refer the readers to this link\footnote{\url{https://www.ece.mcmaster.ca/~xwu/interp_1.pdf}} for more details. Bicubic interpolation yields better performance than the previous two algorithms, however, it also needs huge time to solve the linear equations to find optimal interpolated values. We omit other interpolations methods as they are more complex that these three methods. 

\textbf{\emph{YUV color mode.}} As a matter of fact, there are tons of encoding standards for YUV color space. The formats described here all use 8 bits per pixel location to encode the Y channel (also called the luma channel), and use 8 bits per sample to encode each U or V chroma sample. However, most YUV formats use fewer than 24 bits per pixel on average, because they contain fewer samples of U and V than of Y. The full-size YUV (32 bits per pixel) is represented as 4:4:4, which means no downsampling of chroma channels. Following BT.601~\cite{rec1993bt}, converting RGB to YUV 4:4:4 can be formulated by 

\begin{equation}
\resizebox{0.9\linewidth}{!}{$
\left\{
\begin{aligned}
Y & = \mathrm{round}( 0.256788 \times R + 0.504129 \times G + 0.097906 \times B) +  16 \\ 
U & = \mathrm{round}(-0.148223 \times R - 0.290993 \times G + 0.439216 \times B) + 128 \\
V & = \mathrm{round}( 0.439216 \times R - 0.367788 \times G - 0.071427 \times B) + 128
\end{aligned}
\right..$}
\end{equation}



Here, we can derive an inverse transform from YUV to RGB, 

\begin{equation}
\resizebox{0.9\linewidth}{!}{$
\left\{
\begin{aligned}
R & = \mathrm{clip}(\mathrm{round}( 1.164383 * C                   + 1.596027 * E  ) ) \\
G & = \mathrm{clip}(\mathrm{round}( 1.164383 * C - (0.391762 * D) - (0.812968 * E) ) ) \\
B & = \mathrm{clip}(\mathrm{round}( 1.164383 * C +  2.017232 * D                   ) ) 
\end{aligned}
\right., \text{where  } 
\left\{
\begin{aligned}
C & = Y - 16 \\ 
D & = U - 128 \\ 
E & = V - 128 \\
\end{aligned}
\right..
\label{eq_yuv2rgb}
$}
\end{equation}

Here, $\mathrm{clip}(\cdot)$ denotes clipping to a range of $[0, 255]$. In some implementation~\cite{wiki_yuv}, \autoref{eq_yuv2rgb} can be approximated by:
\begin{equation}
\resizebox{0.9\linewidth}{!}{$
\left\{
\begin{aligned}
R & = \mathrm{clip}(( 298 * C           + 409 * E + 128) >> 8) \\
G & = \mathrm{clip}(( 298 * C - 100 * D - 208 * E + 128) >> 8) \\
B & = \mathrm{clip}(( 298 * C + 516 * D           + 128) >> 8)
\end{aligned}
\right..
$}
\end{equation}
As we could see, the conversion cannot be lossless with the existence of rounding and clipping operations, which could be generally summarized to \emph{quantization-dequantization} conversion. In addition, usually, the hardware supports YUV 4:2:0 rather than 4:4:4, making the conversion to RGB more unstable cause YUV 4:2:0 should be transformed to YUV 4:4:4 and then transformed to RGB format~\cite{wood2005rec}.

\textbf{\emph{Ceiling mode.}} For pooling layers, the output shape of the feature map is calculated by
\begin{equation}
    O = \left\lfloor \frac{W-K+2P}{S} \right\rfloor + 1,
    \label{eq_ceilmode}
\end{equation}
where $W$ is the width (we assume the feature map is square), $K$ is the kernel size, $P$ is the padding size, and $S$ is the stride of pooling layers. The above equation uses floor operation $\lfloor \cdot \rfloor$ to compute the size of the output feature while we can use ceiling operation $\lceil \cdot \rceil$ operation in ceiling mode. Therefore, the border of the output feature is dependent on the ceiling mode. 

\textbf{\emph{Data Precision.}} We here discuss two types of precision: FP16 and INT8. The FP16 still uses floating-point numbers with less bitwidth. According to IEEE 754, the FP32 format uses 1 bit for sign, 8 bits for the exponent, and the rest 23 bits for fraction, while the FP16 uses 1 bit for sign, 5 bits for the exponent, and 10 bits for fraction. Normally, converting FP32 to FP16 only causes a negligible error, as shown in our experiments. For INT8, this is usually done by quantization and de-quantization functions:
\begin{align}
    \bar{\rmX} &= \mathrm{clip}\left(\lfloor\frac{\rmX}{s}\rceil +z, N_{min}, N_{max}\right) \\
    \hat{\rmX} &= s * (\bar{\rmX} - z),
\end{align}
where $\lfloor\cdot\rceil$ is the rounding-to-nearest function. $N_{min}, N_{max}$ are the range of integers that can be represented. For INT8, $N_{min}=-128$ and $N_{max}=127$. $s\in\mathcal{R}$ and $z\in\mathcal{Z}$ are the scale and zero point parameters to fit the original FP32 tensor's range. For more details of quantization, readers are recommended to~\cite{li2021mqbench}.

\textbf{\emph{Post-processing. }}
For object detection, the post-processing involves multiple operations: 1. calculate the anchors, 2. get the offsets for anchors from the predicted outputs, 3. calculate the final bounding box. Some details of these operations are easy to bring the noise. Some details of these operations are easy to cause noise. The following code shows an example procedure for post-processing. For different hardware implementations, the $\mathbf{ALIGNED\_FLAG.offset}$ in the code often has different values of 0 or 1. This minor difference will bring a perturbation to the final accuracy performance. Besides, other operations like the rounding from float-point output to integer coordinate or the precision of exponential also need to be treated carefully.

\definecolor{codeblue}{rgb}{0.25,0.5,0.5}
\definecolor{codekw}{rgb}{0.85, 0.18, 0.50}
\begin{lstlisting}[language=python,basicstyle=\ttfamily, breaklines=true]
# anchors from xyxy format to xywh format
ctr_x, ctr_y, widths, heights = xyxy2xywh(boxes)

# normalize the offsets predicted from the neural network
means = offset.new_tensor(means).view(1, -1).repeat(1, offset.size(-1) // 4)
stds = offset.new_tensor(stds).view(1, -1).repeat(1, offset.size(-1) // 4)
offset = offset * stds + means

# calculate the delta of x, y, w and h
wx, wy, ww, wh = weights
dx = offset[:, 0::4] / wx
dy = offset[:, 1::4] / wy
dw = offset[:, 2::4] / ww
dh = offset[:, 3::4] / wh

dw = torch.clamp(dw, max=np.log(1000. / 16.))
dh = torch.clamp(dh, max=np.log(1000. / 16.))

# calculate the predicted coordinate of center point, 
# and the height & weight of bbox
pred_ctr_x = dx * widths[:, None] + ctr_x[:, None]
pred_ctr_y = dy * heights[:, None] + ctr_y[:, None]
pred_w = torch.exp(dw) * widths[:, None]
pred_h = torch.exp(dh) * heights[:, None]

# calculate the final bbox
pred_boxes = offset.new_zeros(offset.shape)
# x1
pred_boxes[:, 0::4] = pred_ctr_x - 0.5 * pred_w
# y1
pred_boxes[:, 1::4] = pred_ctr_y - 0.5 * pred_h
# x2
pred_boxes[:, 2::4] = pred_ctr_x + 0.5 * pred_w - ALIGNED_FLAG.offset
# y1
pred_boxes[:, 3::4] = pred_ctr_y + 0.5 * pred_h - ALIGNED_FLAG.offset
\end{lstlisting}

\section{Dose Learning-Based Decoder Improve Model Robustness against SysNoise?}
\label{learningbaseddecoder}
Different from the traditional image encoding/decoding method some work uses a learning-based image codec to minimize the gap between the original image and the encoded image. \cite{learningbased} introduce a learning-based image compression method, which achieves about 32.625dB for the CLIC2020 validation dataset. To explore whether the learning-based method can improve the model's robustness on SysNoise, we carried out experiments on the ImageNet dataset using the decoder trained on the CLIC2020 dataset. We used ResNet-50 as a base model, and compare it with the other 2 commonly used decoder methods in \autoref{tab:learn_based}. We can see that there is no obvious gain in using the learning-based decoder.

\begin{table}[htbp]
\centering
\caption{Compare Performance on Learning-Based Decoder}
\label{tab:learn_based}
\adjustbox{max width=0.48\textwidth}{
\begin{tabular}{l|ccc|cc}
\toprule
\diagbox{Train}{Test} & Pillow & OpenCV & Learning-Based & Mean & Std. \\
\midrule
Pillow  & 76.430  & 76.426 & 75.310 & 76.055 & 6.45E-01          \\
OpenCV  & 76.510   & 76.510  & 75.368   & 76.126 & 6.56E-01   \\
Learning-Based & 75.340  & 76.441 & 76.530 & 76.104 & 6.63E-01  \\
\bottomrule
\end{tabular}
}
\end{table}

\section{Preliminary Results for SysNoise on Text-to-Speech Task}
For evaluating SysNoise on text-to-speech tasks, we use FastSpeech 2 \cite{FastSpeech2} and Tacotron 2 \cite{Tacotron2} these two commonly used models. LJ Speech dataset \cite{ljspeech17}, which contains 13,100 English audio clips (about 24 hours) and corresponding text transcripts, was chosen for the training and testing process. Different from other text-to-speech work using MOS(mean opinion score) to evaluate audio quality, we use MSE(mean square error) since we pay more attention to the difference between the generated audio and the original audio under the influence of SysNoise. The result is shown in \autoref{tab_tts}.
From this result, we can tell that the text-to-speech task has a unique SysNoise when doing STFT(short-time Fourier transform). SysNoise introduced by different operators in STFT can harm the model's performance during model inference.

\begin{table}[htbp]
\centering
\caption{\textbf{Measuring SysNoise on Text-to-Speech Taks.} We record MSE. \emph{The lower} MSE \emph{the better}. }
\adjustbox{max width=0.48\textwidth}{
\begin{tabular}{l c c c c}
\toprule
\textbf{Method} &\multicolumn{2}{c}{\textbf{Precision (FP16/INT8)}} & \multicolumn{1}{c}{\textbf{STFT}} & \multicolumn{1}{c}{\textbf{Combined}} \\
\cmidrule(l{2pt}r{2pt}){2-3} \cmidrule(l{2pt}r{2pt}){4-4} \cmidrule(l{2pt}r{2pt}){5-5}  
& MSE & MSE & MSE & MSE\\
\midrule
\multirow{1}{*}{FastSpeech 2} & 0.82 & 1.41 & 2.14 & 4.12 \\
\multirow{1}{*}{Tacotron 2} & 0.71 & 1.21 & 3.01 & 5.02 \\
\bottomrule
\end{tabular}
}
\label{tab_tts}
\end{table}

\section{Broader Impacts and limitations}
\label{limitations}

Together with existing benchmarks on adversarial and natural noises, we could build a more comprehensive and general understanding and ecosystems for robustness benchmarking involving more perspectives. We hope this benchmark could draw the attention of both algorithm researchers and hardware vendors to this inevitable and urgent-to-solve problem, and open a new research direction for building robust deep learning deployment systems. 

Though having investigated several types of SysNoise in this paper, there may still exist other noises that would cause model performance degeneration during deployment. In the future, we will keep the benchmark growing.


\section{Consistency of Results}
\label{error bar}
To maintain consistency of results, we use following method. (1) Fix in the requirements torch==1.8.1, opencv==4.1.1.26 and Pillow==6.2.1 in our framework. (2) Set torch.backends.cudnn.benchmark=True in the code.
We test the ResNet-18 Model on all kinds of noise multiple times in our framework and observe little different result ($<0.0001\%$) on accuracy. This result also holds for object detection and instance segmentation task. So other factors are less likely to affect the results of the model inference process.

\section{Reproducibility and Run Time}
\label{computing_resources}
We provide the code to run this benchmark on GitHub where everyone can download from freely. As for the setup steps and instructions about our code, we provided them in the README file. The installation instructions are also provided in the README file, users can easily install the required run time environment of this codebase. For some noises that need to be generated on specific hardware and are not easy to reproduce, we provide our own resulting datasets generated on specific hardware, which involve ImageNet validation set and COCO validation set. All these datasets can be freely downloaded on our website.

Since our benchmark experiments need us to train multiple models and evaluate them on different kinds of noises, it needs a large amount of GPU resources. The total cost of our GPU resources to build this benchmark is about 5 GPU years. Most of our experiments are run on  Nvidia Tesla V100 GPU. For one training experiment, we run it on 16 GPUs parallel. For inference experiments, we run it on 4 GPUs parallel.

For other users who just want to test their trained model with our framework, the GPU time they require will be greatly reduced. In most cases it only takes 10 to 40 minutes of GPU time to test the effect of one noise on one model, depending on the GPU type they are using.

\section{Future Work}
Based on the research conducted in this paper, our future work will focus on extending the SysNoise to other fields such as speech and audio. We will explore how SysNoise occurs in the different steps of the ML pipeline and benchmark it. We will keep updating our website and the final results will release on it at \url{https://modeltc.github.io/systemnoise_web}

\section{License}
\label{license}
Our code is released under Apache License 2.0. Most model architectures are added to the code with the license chosen by the original author. The ImageNet-1K, COCO, and CitySpace datasets we use are downloaded from the official release. Some system noise datasets we generated from the original dataset follow the license of its original dataset.


\end{document}